\documentclass[11pt,a4paper]{article}

\usepackage[utf8]{inputenc}
\usepackage{multicol}
\usepackage{multirow}

\usepackage{floatpag}
\usepackage{wrapfig, floatrow}
\usepackage[font=small]{caption}
\usepackage{float}
\usepackage[round,authoryear]{natbib}
\usepackage[nottoc,notlot,notlof]{tocbibind}
\usepackage{afterpage}

\usepackage{bm}
\usepackage{amsmath}

\usepackage[normalem]{ulem} 

\usepackage{boxedminipage}

\usepackage[textwidth=1in]{todonotes}

\usepackage{enumitem}
\usepackage[symbol]{footmisc}

\usepackage{textcomp}
\usepackage{listings}

\usepackage{color}
\definecolor{codegreen}{rgb}{0,0.5,0}
\definecolor{codeblue}{rgb}{0,0,0.9}
\definecolor{codegray}{rgb}{0.5,0.5,0.5}
\definecolor{codepurple}{rgb}{0.58,0,0.82}
\definecolor{backcolour}{rgb}{0.95,0.95,0.92}
\definecolor{backcolour2}{rgb}{0.9,0.9,0.9}
\definecolor{codered}{rgb}{0.5,0,0}
\definecolor{textcodered}{rgb}{0.4,0,0}
\definecolor{palegray}{rgb}{0.98,0.98,0.99}

\lstdefinestyle{mystyle}{
    backgroundcolor=\color{backcolour},   
    commentstyle=\color{codered},
    keywordstyle=\color{codeblue},
    numberstyle=\tiny\color{codegray},
    stringstyle=\color{codegreen},
    breakatwhitespace=false,         
    breaklines=true,                 
    captionpos=b,                    
    keepspaces=true,                 
    numbersep=5pt,                  
    showspaces=false,                
    showstringspaces=false,
    showtabs=false,                  
    tabsize=2,
    basicstyle=\ttfamily\footnotesize,
    xleftmargin=2pt, 
    framesep=2pt, 
    frame=l, 
    framerule=0pt
}

\newcommand{\mono}[1]{\texttt{\color{textcodered}#1}}

\newcommand{\dmcontrol}{\mono{dm\_control} }
\newcommand{\dmenv}{\mono{dm\_env} }
\newcommand{\mujoco}{\textsf{MuJoCo} }
\newcommand{\numpy}{\textsf{NumPy} }

\usepackage{ifpdf}

\usepackage{fancyhdr}   
\usepackage{makeidx}

\usepackage{lipsum}
\usepackage{verbatim}
\usepackage{lmodern}

\usepackage[scale=.94]{nimbusmono} 

\usepackage[T1]{fontenc}

\usepackage{booktabs} 

\usepackage{float}
\newfloat{code}{thp}{lop}
\floatname{code}{Code example}

\usepackage{rotating}
\usepackage{blindtext}
\usepackage[includemp=false, marginparwidth=1.8cm, hmargin=3.5cm, vmargin=3cm]{geometry}
\ifnum\pdfoutput>0			
\usepackage[pdftex, 			
pagebackref = false, 			
bookmarks = true, 			
bookmarksnumbered = false, 	
hyperindex = true,			
linktocpage = true,			
pdfpagemode = UseNone, 		
pdfstartview = Fit, 			
pdfpagelayout = OneColumn,
colorlinks = true, 			
urlcolor = blue, 				
citecolor = blue,
linkcolor = blue,
]{hyperref} 					
\fi

\usepackage{tikz}
\usetikzlibrary{shapes,arrows}
\tikzstyle{decision} = [diamond, draw, fill=white, text badly centered, inner sep=0.1em, aspect=4.5]
\tikzstyle{hook} = [rectangle, draw, fill=backcolour, text centered, rounded corners, minimum height=5ex, inner sep=1em]
\tikzstyle{event} = [draw, ellipse, fill=white, text badly centered, minimum height=5ex, inner sep=0.5em]
\tikzstyle{line} = [draw, -latex']

\pgfdeclarelayer{background}
\pgfsetlayers{background,main}

\ifpdf
\DeclareGraphicsExtensions{.png, .pdf, .jpg, .tif}
\else
\DeclareGraphicsExtensions{.eps, .jpg}
\fi

\usepackage{amsmath, amsthm}
\usepackage{amsfonts}
\usepackage{amssymb}

\defcitealias{cmu_mocap}{CMU Motion Capture Database}

\definecolor{lightgrey}{rgb}{.5,.7,.6}

\newcommand{\myfigure}[2]{
\begin{figure}[H]
\floatbox[{\capbeside\thisfloatsetup{capbesideposition={right,top},
capbesidewidth=0.79\textwidth}}]{figure}[\FBwidth]
{\caption*{#1}}
{\includegraphics[width=2.5cm]{#2}}
\end{figure}
\vspace{-.5cm}
}

\graphicspath{{figures/}}

\title{{\vspace{-1cm}\bf \Huge \texttt{dm\_control}\huge: Software and Tasks\\for Continuous Control}}

\author{ 
\renewcommand*{\thefootnote}{\fnsymbol{footnote}}
\setcounter{footnote}{1}
\small Yuval Tassa\textsuperscript{\textdagger}, Saran Tunyasuvunakool\textsuperscript{\textdagger}, Alistair Muldal\textsuperscript{\textdagger}\\
\small Yotam Doron, Piotr Trochim, Siqi Liu, Steven Bohez, Josh Merel\textsuperscript{\textdagger},\\
\small Tom Erez, Timothy Lillicrap, Nicolas Heess\footnote{
Corresponding authors: \texttt{\{tassa, alimuldal, stunya, jsmerel, heess\}@google.com
}}.
}

\date{\small \today}

\begin{document}

\maketitle
\vspace{-.6cm}

\begin{abstract}
\noindent 
The \texttt{dm\_control} software package is a collection of Python libraries and task suites for reinforcement learning agents in an articulated-body simulation. A \textsf{MuJoCo} wrapper provides convenient bindings to functions and data structures. The \textsf{PyMJCF} and \textsf{Composer} libraries enable procedural model manipulation and task authoring. The \textsf{Control Suite} is a fixed set of tasks with standardised structure, intended to serve as performance benchmarks. The \textsf{Locomotion} framework provides high-level abstractions and examples of locomotion tasks. A set of configurable \textsf{manipulation} tasks with a robot arm and snap-together bricks is also included.
\vspace{0cm}
\center{\texttt{dm\_control} is publicly available at \href{https://www.github.com/deepmind/dm_control}{\textsf{github.com/deepmind/dm\_control}}.}
\end{abstract}




\renewcommand*{\thefootnote}{\arabic{footnote}}
\setcounter{footnote}{0}

\begin{figure*}[ht]
\begin{minipage}[c]{1\textwidth}
\def\mywidth{1.95cm}
\def\myhsep{-2mm}
\includegraphics[width=\mywidth]{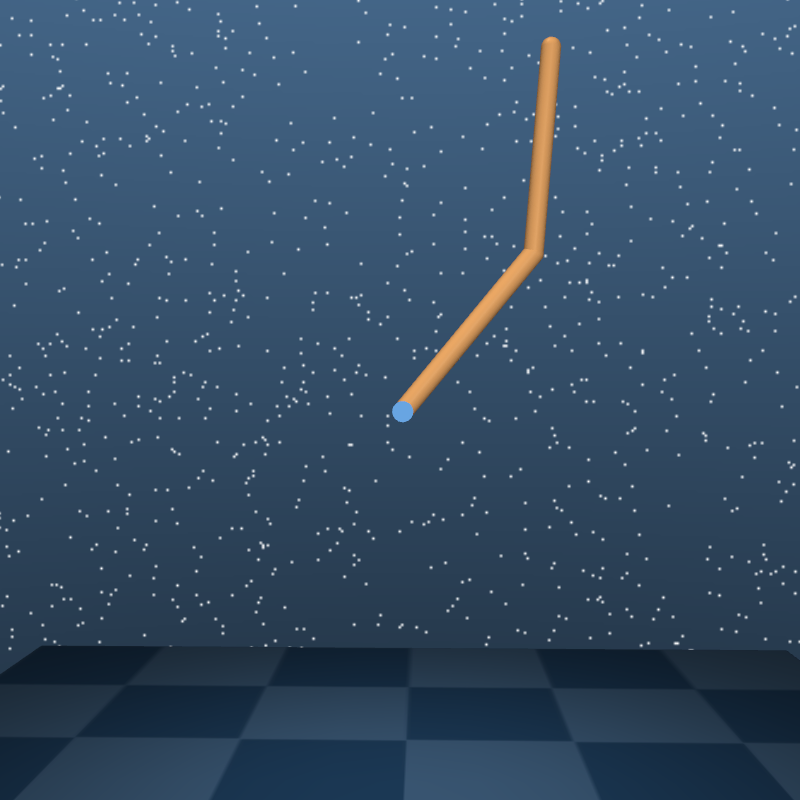}
\hspace{\myhsep}
\includegraphics[width=\mywidth]{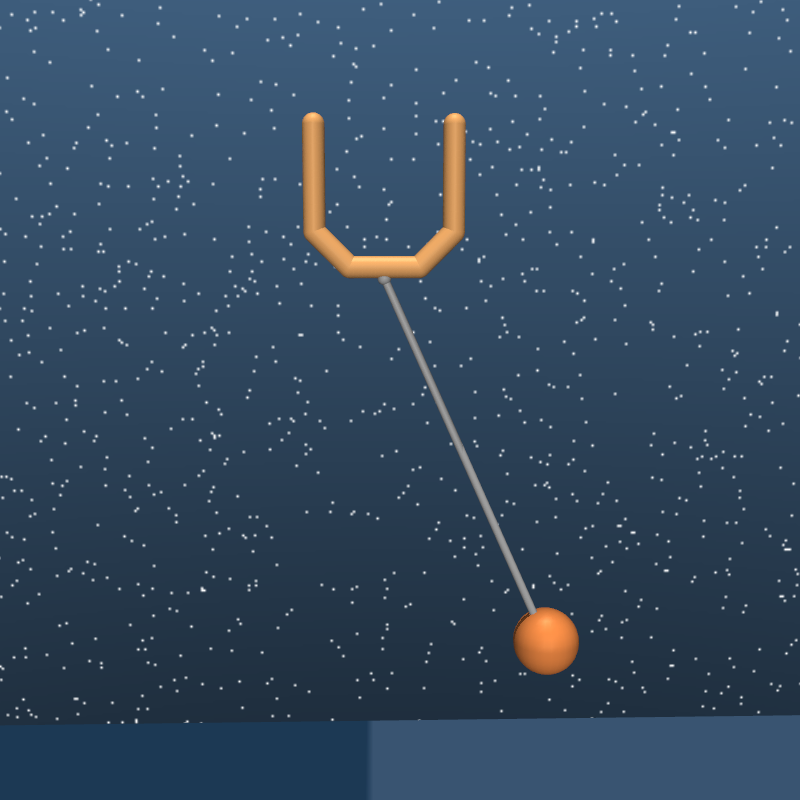}
\hspace{\myhsep}
\includegraphics[width=\mywidth]{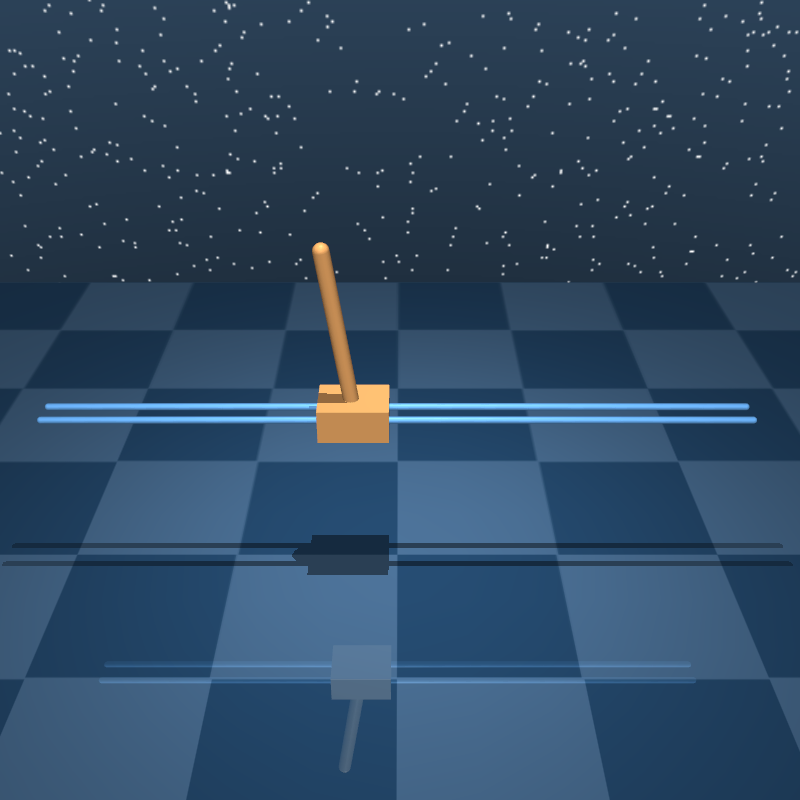}
\hspace{\myhsep}
\includegraphics[width=\mywidth]{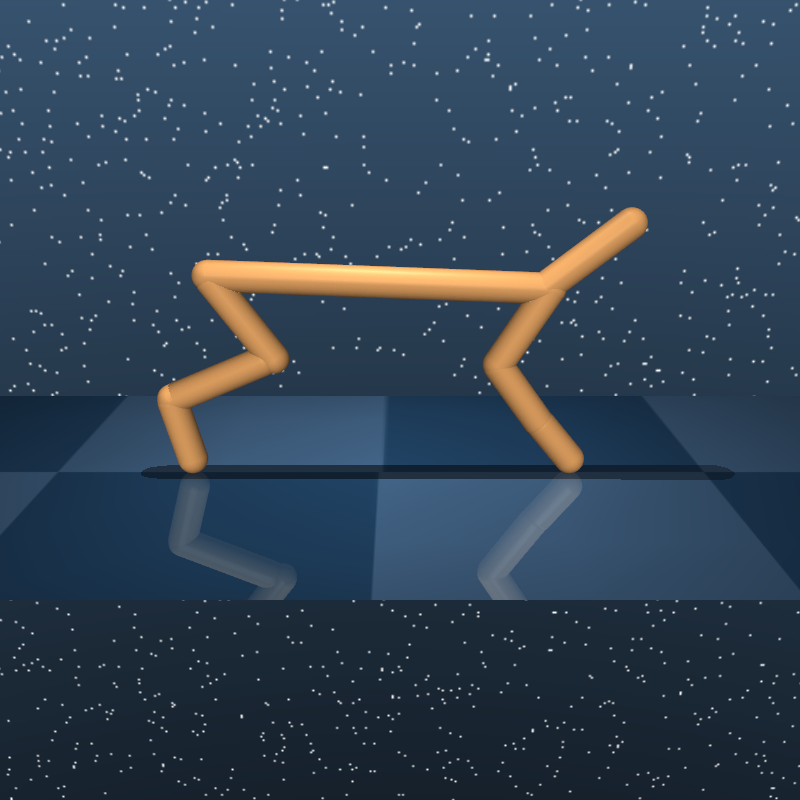}
\hspace{\myhsep}
\includegraphics[width=\mywidth]{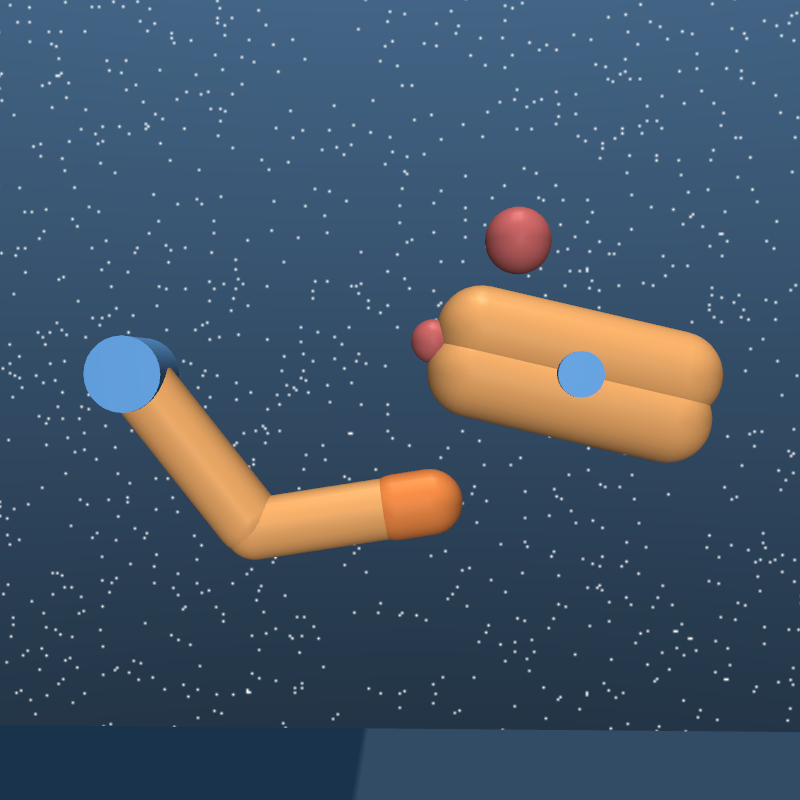}
\hspace{\myhsep}
\includegraphics[width=\mywidth]{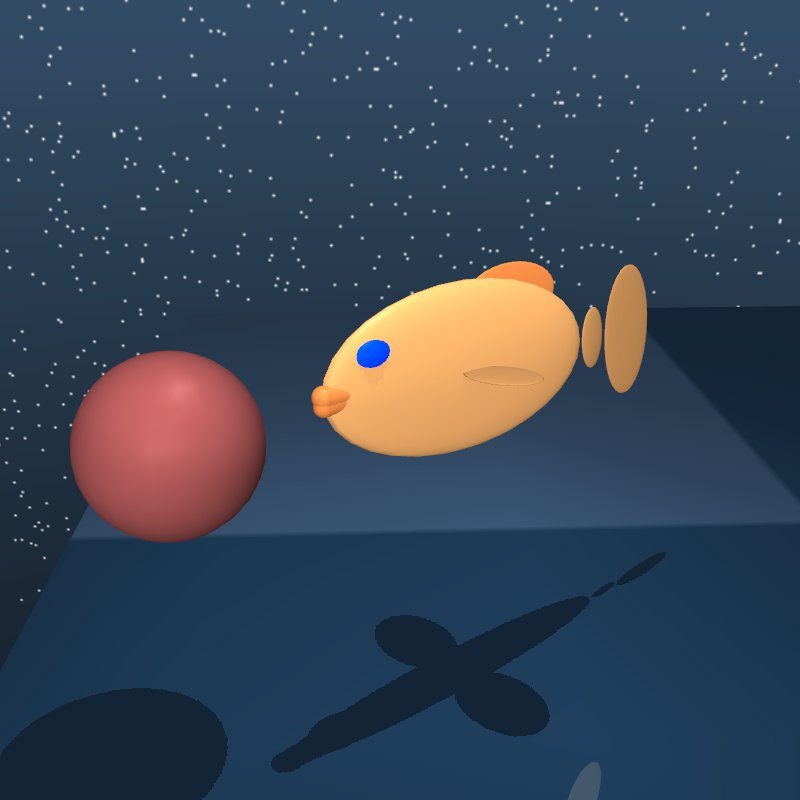}
\hspace{\myhsep}
\includegraphics[width=\mywidth]{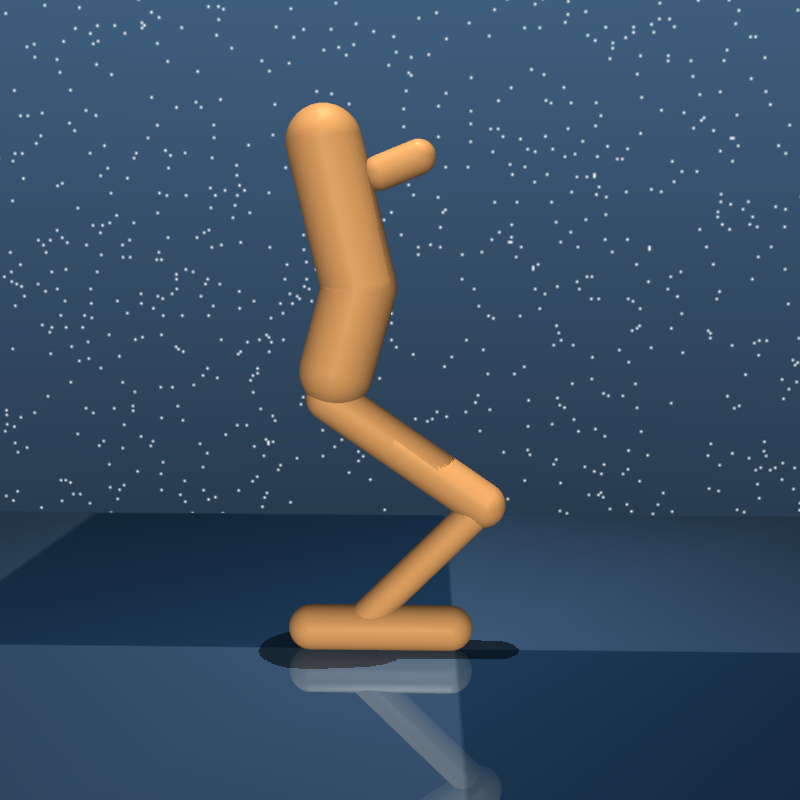}\\
\includegraphics[width=\mywidth]{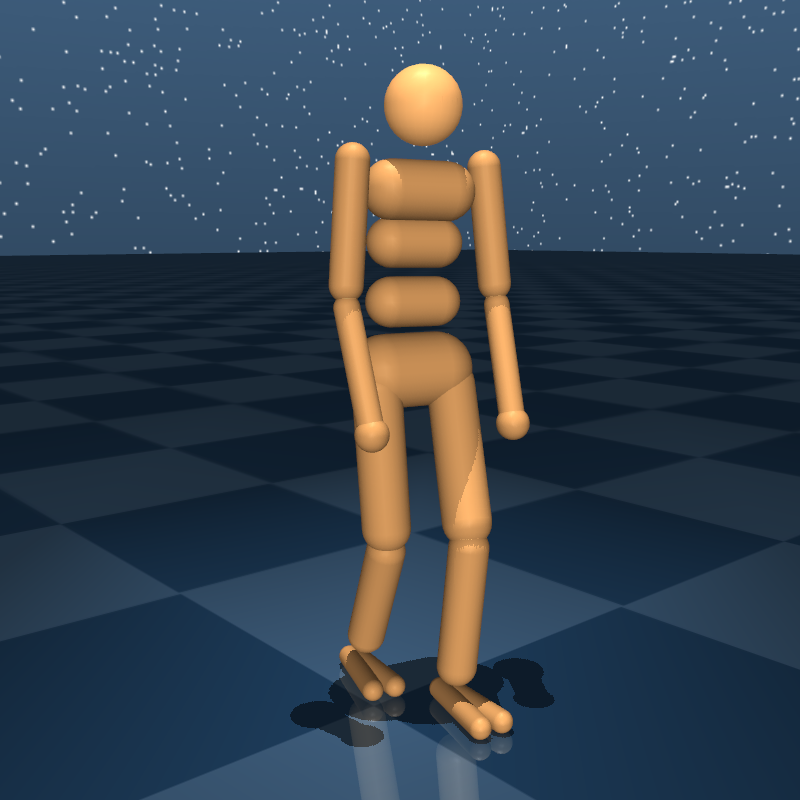}
\hspace{\myhsep}
\includegraphics[width=\mywidth]{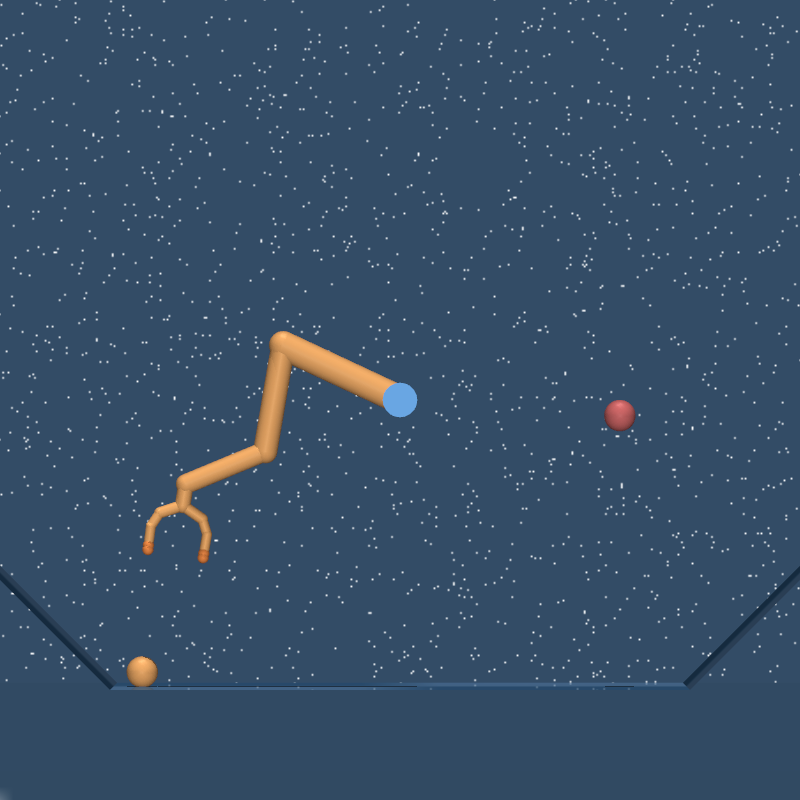}
\hspace{\myhsep}
\includegraphics[width=\mywidth]{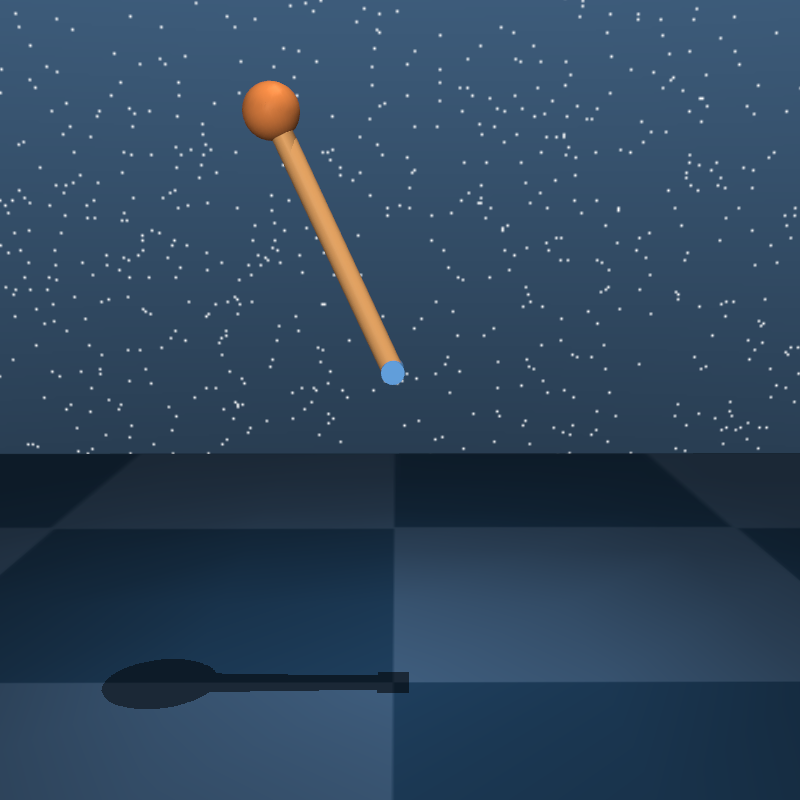}
\hspace{\myhsep}
\includegraphics[width=\mywidth]{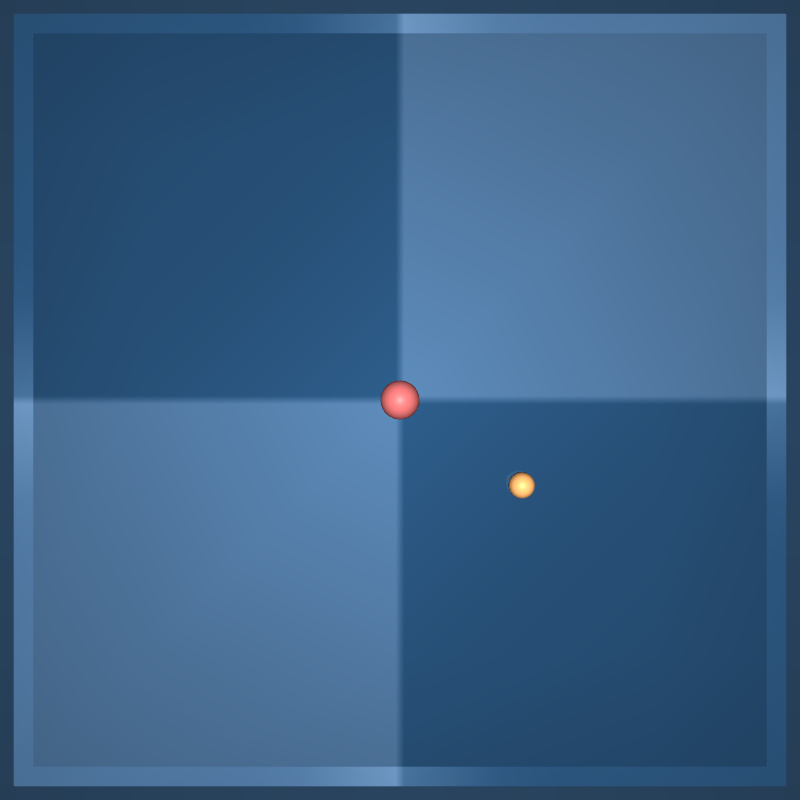}
\hspace{\myhsep}
\includegraphics[width=\mywidth]{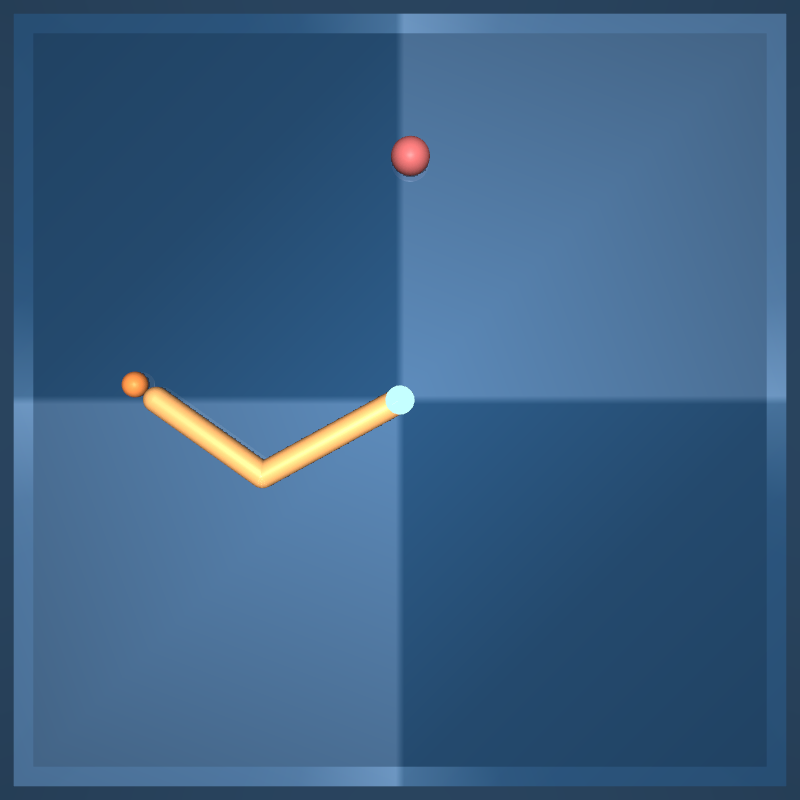}
\hspace{\myhsep}
\includegraphics[width=\mywidth]{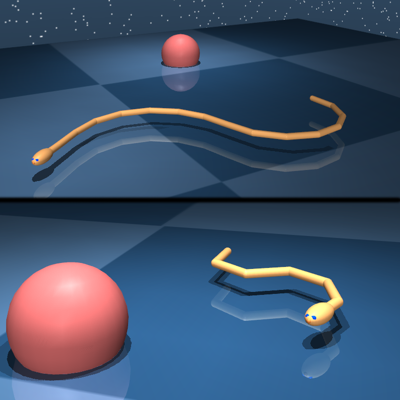}
\hspace{\myhsep}
\includegraphics[width=\mywidth]{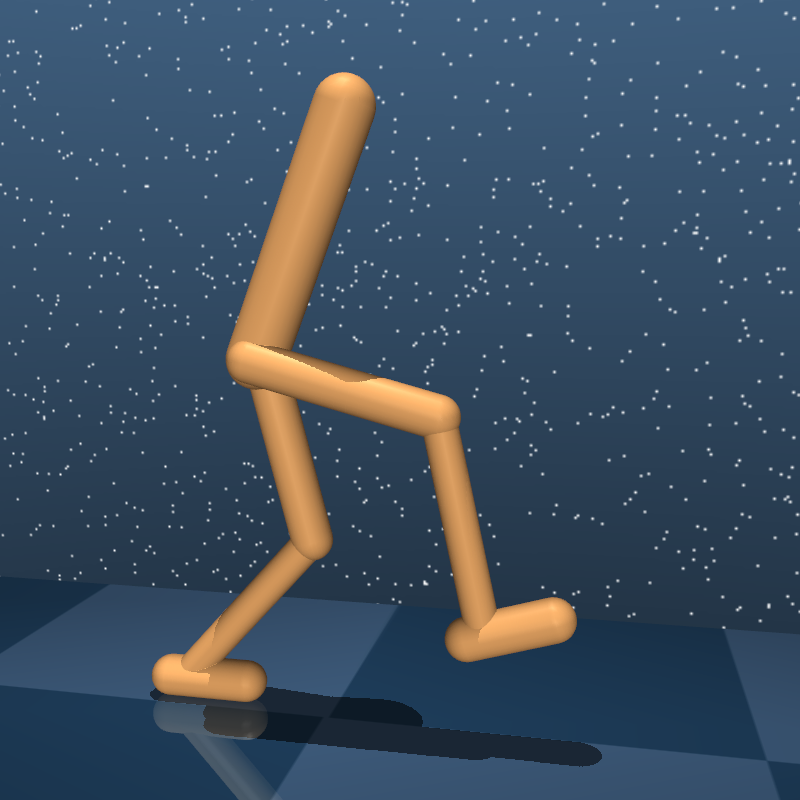}

\caption{The \textsf{Control Suite} benchmarking domains, described in Section \ref{sec:controlsuite} (\href{https://youtu.be/rAai4QzcYbs}{see video}).}
\label{fig:benchmarking}
\end{minipage}
\end{figure*}
\begin{figure*}[h]
\centering
\begin{minipage}[c]{1\textwidth}
\includegraphics[height=0.25\textwidth]{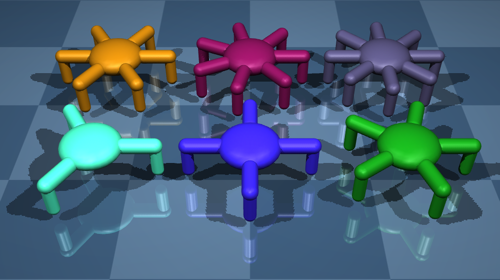}
\hspace{.001\textwidth}
\includegraphics[height=0.25\textwidth]{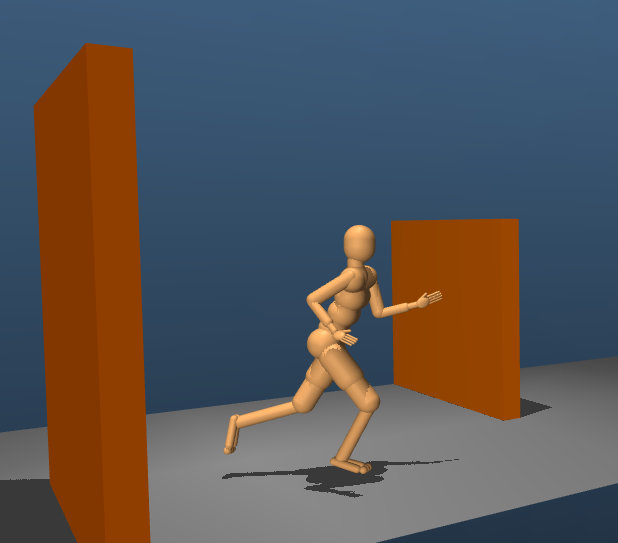}
\hspace{.001\textwidth}
\includegraphics[height=0.25\textwidth]{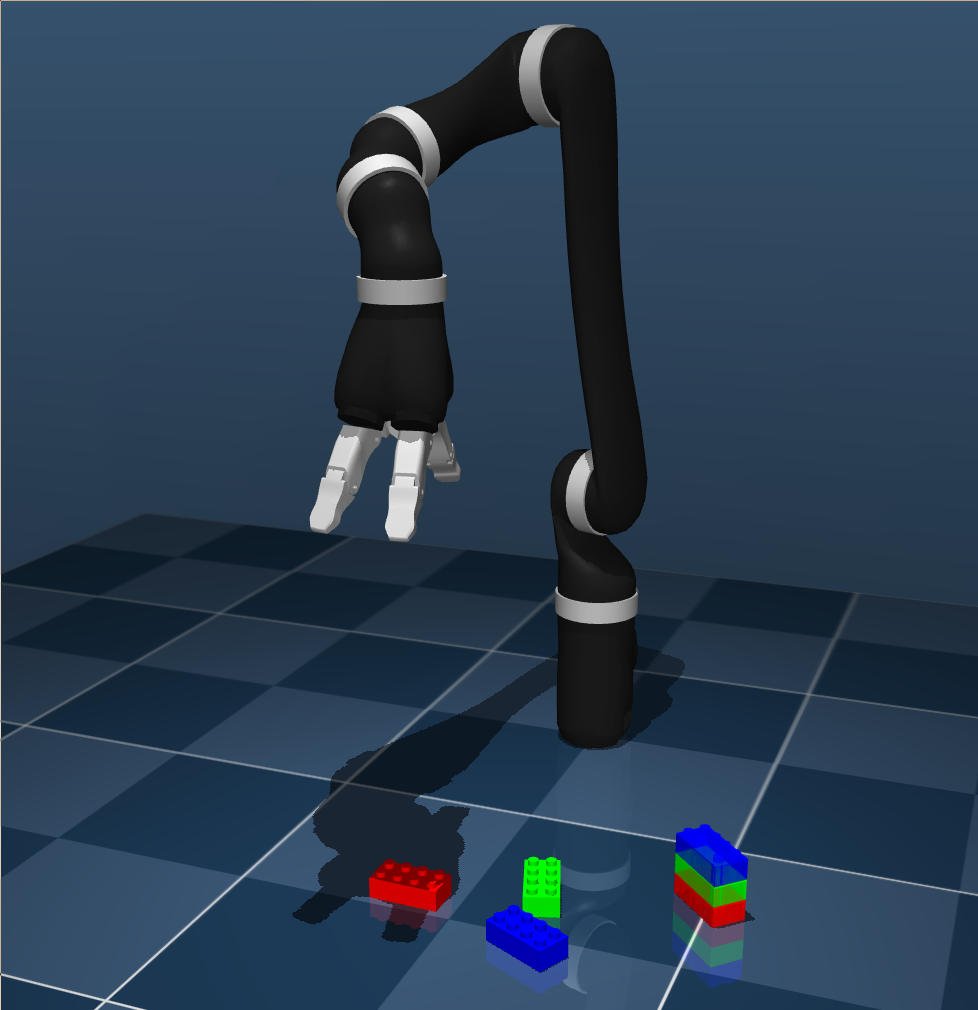}
\caption{\footnotesize Procedural domains built with the \textsf{PyMJCF} (Sec.\ \ref{sec:pymjcf}) and \textsf{Composer} (Sec.\ \ref{sec:composer}) task-authoring libraries. \textit{Left}: Multi-legged creatures from the tutorial in Sec.\ \ref{sec:pymjcftutorial}. \textit{Middle}: The ``run through corridor'' example task from Sec.\ \ref{sec:locomotion_walls}. \textit{Right}: The ``stack 3 bricks'' example task from Sec.\ \ref{sec:manipulation}.}
\end{minipage}
\end{figure*}


\newpage
\tableofcontents
\newpage

\section{Introduction}

Controlling the physical world is an integral part and arguably a prerequisite of general intelligence \citep{wolpert2003}. Indeed, the only known example of general-purpose intelligence emerged in primates whose behavioural niche was already contingent on two-handed manipulation for millions of years.

Unlike board games, language and other symbolic domains, physical tasks are fundamentally \textit{continuous} in state, time and action. 
Physical dynamics are subject to second-order equations of motion -- the underlying state is composed of positions and velocities. Sensory signals (i.e.\ observations) carry meaningful physical units and vary over corresponding timescales. These properties, along with their prevalence and importance, make control problems a unique subset of general Markov Decision Processes. The most familiar physical control tasks have a fixed subset of degrees of freedom (the \emph{body}) that are directly actuated, while the rest are unactuated (the \emph{environment}). Such \emph{embodied} tasks are the focus of \mono{dm\_control}.

\subsection{Software for research}
The \dmcontrol package was designed by DeepMind scientists and engineers to facilitate their own continuous control and robotics needs, and is therefore well-suited for research. It is written in \textsf{Python}, exploiting the agile workflow of a dynamic language, while relying on the \textsf{C}-based \mujoco physics library, a fast and accurate simulator \citep{erez2015simulation}, itself designed to facilitate research \citep{todorov2012mujoco}.
It is composed of the following modules:
\begin{itemize}
\item The \mono{ctypes}-based \mujoco wrapper (Sec.~\ref{sec:mujoco}) provides full access to the simulator, conveniently exposing quantities with named indexing. A Python-based interactive visualiser (Sec.~\ref{sec:viewer}) allows the user to examine and perturb scene elements with a mouse.
\item The \textsf{PyMJCF} library (Sec.~\ref{sec:pymjcf}) can procedurally assemble model elements and allows the user to configure or randomise parameters and initial states.
\item An environment API that exposes actions, observations, rewards and terminations in a consistent yet flexible manner (Sec.~\ref{sec:rl}).
\item Finally, we combine the above functionality in the high-level task-definition framework  \textsf{Composer} (Sec.~\ref{sec:composer}). Amongst other things it provides a Model Variation module (Sec.~\ref{sec:variation}) for policy robustification, and an Observable module for delayed, corrupted, and stacked sensor data (Sec.~\ref{sec:observable}).
\end{itemize}
\dmcontrol has been used extensively in DeepMind, serving as a fundamental component of continuous control research. See \href{https://www.youtube.com/watch?v=CMjoiU482Jk}{\textsf{youtu.be/CMjoiU482Jk}} for a montage of clips from selected publications.

\subsection{Tasks}
Recent years have seen rapid progress in the application of Reinforcement Learning (RL) to difficult problem domains such as video games \citep{Mnih2015}. The Arcade Learning Environment (\textsf{ALE}, \citealt{bellemare2012arcade}) was and continues to be a vital facilitator of these developments, providing a set of standard benchmarks for evaluating and comparing learning algorithms. Similarly, it could be argued that control and robotics require well-designed task suites as a standardised playing field, where different approaches can compete and new ones can emerge.

The OpenAI \textsf{Gym}~\citep{brockman2016gym} includes a set of continuous control domains that have become a popular benchmark in continuous RL~\citep{duan2016benchmarking, 2017deepRLmatters}. More recent task suites such as \textsf{Meta-world}~\citep{yu2019metaworld}, \textsf{SURREAL}~\citep{corl2018surreal}, \textsf{RLbench}~\citep{james2019rlbench} and \textsf{IKEA}~\citep{lee2019ikea}, have been published in an attempt to satisfy the demand for tasks suites that facilitate the study of algorithms related to multi-scale control, multi-task transfer, and meta learning. \dmcontrol includes its own sets of control tasks, in three categories:

\begin{description}[,leftmargin=.6cm,itemindent=-.6cm]
\item{\textbf{\textsf{Control Suite}}}

The DeepMind \textsf{Control Suite} (Section \ref{sec:controlsuite}), first introduced in \citep{deepmindcontrolsuite2018}, built directly with the \mujoco wrapper, provides a set of standard benchmarks for continuous control problems. The unified reward structure offers interpretable learning curves and aggregated suite-wide performance measures. Furthermore, we emphasise high-quality, well-documented code using uniform design patterns, offering a readable, transparent and easily extensible codebase.
\item{\textbf{\textsf{Locomotion}}}

The \textsf{Locomotion} framework (Section \ref{sec:locomotion}) was inspired by our work in \cite{heess2017emergence}. It is designed to facilitate the implementation of a wide range of locomotion tasks for RL algorithms by introducing self-contained, reusable components which compose into different task variants. The \textsf{Locomotion} framework has enabled a number of research efforts including \cite{merel2017learning}, \cite{merel2018neural}, \cite{merel2018hierarchical} and more recently has been employed to support Multi-Agent domains in \cite{liu2019emergent}, \cite{sunehag2019} and \cite{banarse2019body}.

\item{\textbf{\textsf{Manipulation}}}

We also provide examples of constructing robotic manipulation tasks (Sec. \ref{sec:manipulation}). These tasks involve grabbing and manipulating objects with a 3D robotic arm. The set of tasks includes examples of reaching, placing, stacking, throwing, assembly and disassembly.
The tasks are designed to be solved using a simulated 6 degree-of-freedom robotic arm based on the \textsf{Kinova Jaco}~\citep{2017jaco}, though their modular design permit the use of other arms with minimal changes. 
These tasks make use of reusable components such as bricks that snap together, and provide examples of reward functions for manipulation.
Tasks can be run using vision, low-level features, or combinations of both.

\end{description}

\newpage
\part{Software Infrastructure}
Sections \ref{sec:mujoco}, \ref{sec:pymjcf}, \ref{sec:rl} and \ref{sec:composer} include code snippets showing how to use \dmcontrol software. These snippets are collated in a \textsf{Google Colab} notebook:
\begin{center}
\href{https://colab.sandbox.google.com/github/deepmind/dm_control/blob/master/tutorial.ipynb}{\textsf{github.com/deepmind/dm\_control/tutorial.ipynb}}
\end{center}
\section{\textsf{MuJoCo} Python interface}
\label{sec:mujoco}
The \mono{mujoco} module provides a general-purpose wrapper of the \textsf{MuJoCo} engine, using \textsf{Python}'s \href{https://docs.python.org/3/library/ctypes.html}{\textsf{ctypes}} library to auto-generate bindings to \textsf{MuJoCo} structs, enums and API functions. We provide a brief introductory overview which assumes familiarity with \textsf{Python}; see in-code documentation for more detail.
\subsection*{\textsf{MuJoCo} physics}
\textsf{MuJoCo}~\citep{todorov2012mujoco} is a fast, reduced-coordinate, continuous-time physics engine. It compares favourably to other popular simulators~\citep{erez2015simulation}, especially for articulated, low-to-medium  degree-of-freedom regimes ($\lessapprox 100$) in the presence of contacts. The~\href{http://mujoco.org/book/modeling.html}{MJCF} model definition format and reconfigurable computation pipeline have made \textsf{MuJoCo} a popular choice for robotics and reinforcement learning research (e.g.\ \citealt{schulman2015trust,heess2015learning,lillicrap2015continuous,duan2016benchmarking}).
\subsection{The \texttt{Physics} class}
The \mono{Physics} class encapsulates \textsf{MuJoCo}'s most commonly used  functionality.
\begin{description}[,leftmargin=0cm,itemindent=0cm]
\item{\textbf{Loading an MJCF model}}

The \mono{Physics.from\_xml\_string()} constructor loads an MJCF model and returns a \mono{Physics} instance:
\begin{lstlisting}[language=Python]
from dm_control import mujoco
simple_MJCF = """
<mujoco>
  <worldbody>
    <light name="top" pos="0 0 1"/>
    <body name="box_and_sphere" euler="0 0 -30">
      <joint name="swing" type="hinge" axis="1 -1 0" pos="-.2 -.2 -.2"/>
      <geom name="red_box" type="box" size=".2 .2 .2" rgba="1 0 0 1"/>
      <geom name="green_sphere" pos=".2 .2 .2" size=".1" rgba="0 1 0 1"/>
    </body>
  </worldbody>
</mujoco>
"""
physics = mujoco.Physics.from_xml_string(simple_MJCF)
\end{lstlisting}

\item{\textbf{Rendering}}

The \mono{Physics.render()} method outputs a \numpy array of pixel values. 
\begin{lstlisting}[language=Python]
pixels = physics.render()
\end{lstlisting}
\vspace{-.2cm}
\begin{figure}[H]
\floatbox[{\capbeside\thisfloatsetup{capbesideposition={right,top},
capbesidewidth=0.65\textwidth}}]{figure}[\FBwidth]
{\caption*{\normalsize Optional arguments to \mono{render()} specify the resolution, camera ID, whether to render RGB, depth or segmentation images, and other visualisation options (e.g. the joint visualisation on the left).
\dmcontrol on Linux supports both OSMesa software rendering and hardware-accelerated rendering using either EGL or GLFW.
The rendering backend can be selected by setting the \mono{MUJOCO\_GL} environment variable to \mono{glfw}, \mono{egl}, or \mono{osmesa}, respectively.}}
{\includegraphics[width=4.2cm]{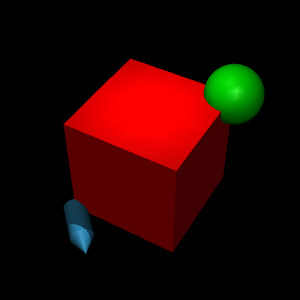}}
\end{figure}

\item{\textbf{\texttt{Physics.model} and \texttt{Physics.data}}}

\textsf{MuJoCo}'s underlying \mono{mjModel} and \mono{mjData} data structures, describing static and dynamic simulation parameters respectively, can be accessed via the \mono{model} and \mono{data} properties of \mono{Physics}. They contain \numpy arrays that have direct, writeable views onto \textsf{MuJoCo}'s internal memory. Because the memory is owned by \textsf{MuJoCo}, an attempt to overwrite an entire array throws an error:
\begin{lstlisting}[language=Python]
# This fails with `AttributeError: can't set attribute`:
physics.data.qpos = np.random.randn(physics.model.nq)
# This succeeds:
physics.data.qpos[:] = np.random.randn(physics.model.nq)
\end{lstlisting}

\item{\textbf{\texttt{mj\_step()} and \texttt{Physics.step()}}}

\textsf{MuJoCo}'s top-level \mono{mj\_step()} function computes the next state --- the joint-space configuration \mono{qpos} and velocity \mono{qvel} --- in two stages. Quantities that depend only on the state are computed in the first stage, \mono{mj\_step1()}, and those that also depend on the control (including forces) are computed in the subsequent \mono{mj\_step2()}. \mono{Physics.step()} calls these sub-functions in reverse order, as follows. Assuming that \mono{mj\_step1()} has already been called, it first completes the computation of the new state with \mono{mj\_step2()}, and then calls \mono{mj\_step1()}, updating the quantities that depend on the state alone. In particular, this means that after a \mono{Physics.step()}, rendered pixels will correspond to the current state, rather than the previous one. Quantities that depend on forces, like accelerometer and touch sensor readings, are still with respect to the last transition.

\item{\textbf{Setting the state with \texttt{reset\_context()}}}

For the above assumption above to hold, \mono{mj\_step1()} must always be called after setting the state. We therefore provide the \mono{Physics.reset\_context()}, within which the state should be set:
\begin{lstlisting}[language=Python]
with physics.reset_context():
   # mj_reset() is called upon entering the context: default state.
   physics.data.qpos[:] = ...  # Set position.
   physics.data.qvel[:] = ...  # Set velocity.
# mj_forward() is called upon exiting the context. Now all derived
# quantities and sensor measurements are up-to-date.
\end{lstlisting}
Note that we call \mono{mj\_forward()} upon exit, which includes \mono{mj\_step1()}, but continues up to the computation of accelerations (but does not increment the state). This is so that force- or acceleration-dependent sensors have sensible values even at the initial state, before any steps have been taken.

\vspace{.3cm}
\item{\textbf{Named indexing}}

Everything in a \mujoco model can be named. It is often more convenient and less error-prone to refer to model elements by name rather than by index. To address this, \mono{Physics.named.model} and \mono{Physics.named.data} provide array-like containers that provide convenient named views. Using our simple model above:
\begin{lstlisting}[language=Python]
print("The `geom_xpos` array:")
print(physics.data.geom_xpos)
print("Is much easier to inspect using `Physics.named`:")
print(physics.named.data.geom_xpos)
\end{lstlisting}
\vspace{-.2cm}
\begin{lstlisting}[backgroundcolor=\color{palegray}]
The `geom_xpos` array:
[[0.         0.         0.        ]
 [0.27320508 0.07320508 0.2       ]]
Is much easier to inspect using `Physics.named`:
                 x         y         z         
0      red_box [ 0         0         0       ]
1 green_sphere [ 0.273     0.0732    0.2     ]
\end{lstlisting}
These containers can be indexed by name for both reading and writing, and support most forms of \numpy indexing:
\begin{lstlisting}[language=Python,morekeywords={*,with}]
with physics.reset_context():
  physics.named.data.qpos['swing'] = np.pi
print(physics.named.data.geom_xpos['green_sphere', ['z']])
\end{lstlisting}
\vspace{-.2cm}
\begin{lstlisting}[backgroundcolor=\color{palegray}]
[-0.6]
\end{lstlisting}
Note that in the example above we use a joint name in order to index into the generalised position array \mono{qpos}. Indexing into a multi-DoF \mono{ball} or \mono{free} joint outputs the appropriate slice. We also provide convenient access to \textsf{MuJoCo}'s \mono{mj\_id2name} and \mono{mj\_name2id}:
\begin{lstlisting}[language=Python]
physics.model.id2name(0, "geom")
\end{lstlisting}
\vspace{-.2cm}
\begin{lstlisting}[backgroundcolor=\color{palegray},upquote=true]
'red_box'
\end{lstlisting}

\end{description}

\subsection{Interactive Viewer}
\label{sec:viewer}
The \mono{viewer} module provides playback and interaction with physical models using mouse input.  This type of visual debugging is often critical for cases when an agent finds an ``exploit'' in the physics. 

\begin{lstlisting}[language=Python]
from dm_control import suite, viewer

environment = suite.load(domain_name="humanoid", task_name="stand")

# Define a uniform random policy.
spec = environment.action_spec()
def random_policy(time_step):
  return np.random.uniform(spec.minimum, spec.maximum, spec.shape)

# Launch the viewer application.
viewer.launch(environment, policy=random_policy)
\end{lstlisting}
See the documentation at \href{https://github.com/deepmind/dm_control/tree/master/dm_control/viewer}{\textsf{dm\_control/tree/master/dm\_control/viewer}} for a screen capture of the \mono{viewer} application. 

\subsection{Wrapper bindings}
The bindings provide easy access to all \textsf{MuJoCo} library functions and enums, automatically converting \numpy arrays to data pointers where appropriate.
\begin{lstlisting}[language=Python,morekeywords={*,as}]
from dm_control.mujoco.wrapper.mjbindings import mjlib
import numpy as np

quat = np.array((.5, .5, .5, .5))
mat = np.zeros(9)
mjlib.mju_quat2Mat(mat, quat)

print("MuJoCo converts this quaternion:")
print(quat)
print("To this rotation matrix:")
print(mat.reshape(3,3))
\end{lstlisting}
\vspace{-.2cm}
\begin{lstlisting}[backgroundcolor=\color{palegray}]
MuJoCo converts this quaternion:
[ 0.5  0.5  0.5  0.5]
To this rotation matrix:
[[ 0.  0.  1.]
 [ 1.  0.  0.]
 [ 0.  1.  0.]]
\end{lstlisting}
Enums are exposed as a submodule:
\begin{lstlisting}[language=Python]
from dm_control.mujoco.wrapper.mjbindings import enums
print(enums.mjtJoint)
\end{lstlisting}
\vspace{-.2cm}
\begin{lstlisting}[backgroundcolor=\color{palegray}]
mjtJoint(mjJNT_FREE=0, mjJNT_BALL=1, mjJNT_SLIDE=2, mjJNT_HINGE=3)
\end{lstlisting}
\section{The \textsf{PyMJCF} library}
\label{sec:pymjcf}
The \textsf{PyMJCF} library provides a Python object model for \textsf{MuJoCo}'s MJCF modelling language, which can describe complex scenes with articulated bodies. The goal of the library is to allow users to interact with and modify MJCF models programmatically using Python, similarly to what the JavaScript DOM does for HTML.

A key feature of the library is the ability to compose multiple MJCF models into a larger one, while automatically maintaining a consistent, collision-free namespace. Additionally, it provides Pythonic access to the underlying \textsf{C} data structures with the \mono{bind()} method of \mono{mjcf.Physics}, a subclass of \mono{Physics} which associates a compiled model with the \textsf{PyMJCF} object tree.

\subsection{\textsf{PyMJCF} Tutorial}
\label{sec:pymjcftutorial}
The following code snippets constitute a tutorial example of a typical use case.

\begin{lstlisting}[language=Python,deletendkeywords={dir,type}]
from dm_control import mjcf

class Leg(object):
  """ A 2-DoF leg with position actuators."""
  def __init__(self, length, rgba):
    self.model = mjcf.RootElement()

    # Defaults:
    self.model.default.joint.damping = 2
    self.model.default.joint.type = 'hinge'
    self.model.default.geom.type = 'capsule'
    self.model.default.geom.rgba = rgba  # Continued below...
    # Thigh:
    self.thigh = self.model.worldbody.add('body')
    self.hip = self.thigh.add('joint', axis=[0, 0, 1])
    self.thigh.add('geom', fromto=[0, 0, 0, length, 0, 0], size=[length/4])

    # Shin:
    self.shin = self.thigh.add('body', pos=[length, 0, 0])
    self.knee = self.shin.add('joint', axis=[0, 1, 0])
    self.shin.add('geom', fromto=[0, 0, 0, 0, 0, -length], size=[length/5])

    # Position actuators:
    self.model.actuator.add('position', joint=self.hip, kp=10)
    self.model.actuator.add('position', joint=self.knee, kp=10)
\end{lstlisting}
The \mono{Leg} class describes an abstract articulated leg, with two joints and corresponding proportional-derivative actuators. Note the following.
\begin{itemize}
\item MJCF attributes correspond directly to arguments of the \mono{add()} method\footnote{The exception is the \mono{class} attribute which is a reserved \textsf{Python} symbol and renamed \mono{dclass}.}.
\item When referencing elements, e.g.\ when specifying the joint to which an actuator is attached in the last two lines above, the MJCF element itself can be used, rather than its name (though a name string is also supported).
\end{itemize}

\begin{lstlisting}[language=Python,deletendkeywords={dir,type}]
BODY_RADIUS = 0.1
BODY_SIZE = (BODY_RADIUS, BODY_RADIUS, BODY_RADIUS / 2)

def make_creature(num_legs):
  """Constructs a creature with `num_legs` legs."""
  rgba = np.random.uniform([0, 0, 0, 1], [1, 1, 1, 1])
  model = mjcf.RootElement()
  model.compiler.angle = 'radian'  # Use radians.

  # Make the torso geom.
  torso = model.worldbody.add(
      'geom', name='torso', type='ellipsoid', size=BODY_SIZE, rgba=rgba)
  
  # Attach legs to equidistant sites on the circumference.
  for i in range(num_legs):
    theta = 2 * i * np.pi / num_legs
    hip_pos = BODY_RADIUS * np.array([np.cos(theta), np.sin(theta), 0])
    hip_site = model.worldbody.add('site', pos=hip_pos, euler=[0, 0, theta])
    leg = Leg(length=BODY_RADIUS, rgba=rgba)
    hip_site.attach(leg.model)

  return model
\end{lstlisting}
The \mono{make\_creature} function uses \textsf{PyMJCF}'s \mono{attach()} method to procedurally attach legs to the torso. Note that both the torso and hip attachment sites are children of the \mono{worldbody}, since their parent body has yet to be instantiated. We will now make an arena with a chequered floor and two lights:
\begin{lstlisting}[language=Python,deletendkeywords={dir,type}]
arena = mjcf.RootElement()
checker = arena.asset.add('texture', type='2d', builtin='checker', width=300,
                          height=300, rgb1=[.2, .3, .4], rgb2=[.3, .4, .5])
grid = arena.asset.add('material', name='grid', texture=checker, 
                       texrepeat=[5,5], reflectance=.2)
arena.worldbody.add('geom', type='plane', size=[2, 2, .1], material=grid)
for x in [-2, 2]:
  arena.worldbody.add('light', pos=[x, -1, 3], dir=[-x, 1, -2])
\end{lstlisting}
Placing several creatures in the arena, arranged in a grid:
\begin{lstlisting}[language=Python,deletendkeywords={dir,type}]
# Instantiate 6 creatures with 3 to 8 legs.
creatures = [make_creature(num_legs=num_legs) for num_legs in (3,4,5,6,7,8)]

# Place them on a grid in the arena.
height = .15
grid = 5 * BODY_RADIUS
xpos, ypos, zpos = np.meshgrid([-grid, 0, grid], [0, grid], [height])
for i, model in enumerate(creatures):
  # Place spawn sites on a grid.
  spawn_pos = (xpos.flat[i], ypos.flat[i], zpos.flat[i])
  spawn_site = arena.worldbody.add('site', pos=spawn_pos, group=3)  
  # Attach to the arena at the spawn sites, with a free joint.
  spawn_site.attach(model).add('freejoint')

# Instantiate the physics and render.
physics = mjcf.Physics.from_mjcf_model(arena)
pixels = physics.render()
\end{lstlisting}
\vspace{-.2cm}
\begin{figure}[H]
\floatbox[{\capbeside\thisfloatsetup{capbesideposition={left,top},
capbesidewidth=0.42\textwidth}}]{figure}[\FBwidth]
{\caption*{\normalsize Multi-legged creatures, ready to roam! Let us inject some controls and watch them move. We will generate a sinusoidal open-loop control signal of fixed frequency and random phase, recording both a video and the horizontal positions of the torso geoms, in order to plot the movement trajectories. }}
{\includegraphics[width=0.55\textwidth]{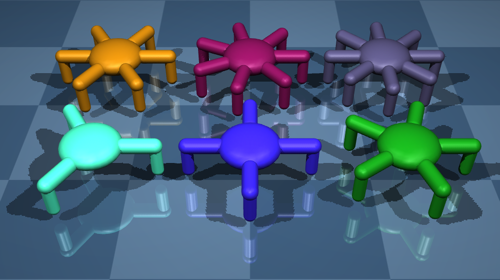}}
\end{figure}
\vspace{-.2cm}
\begin{lstlisting}[language=Python]
duration  = 10  # (Seconds)
framerate = 30  # (Hz)
video = []; pos_x = []; pos_y = []
torsos = []  # List of torso geom elements.
actuators = []  # List of actuator elements.
for creature in creatures:
  torsos.append(creature.find('geom','torso'))
  actuators.extend(creature.find_all('actuator'))

# Control signal frequency, phase, amplitude.
freq = 5
phase = 2 * np.pi * np.random.rand(len(actuators))
amp = 0.9

# Simulate, saving video frames and torso locations.
physics.reset()
while physics.data.time < duration:
  # Inject controls and step the physics.
  physics.bind(actuators).ctrl = amp*np.sin(freq*physics.data.time + phase)
  physics.step()

  # Save torso horizontal positions using bind().
  pos_x.append(physics.bind(torsos).xpos[:, 0].copy())
  pos_y.append(physics.bind(torsos).xpos[:, 1].copy())

  # Save video frames.
  if len(video) < physics.data.time * framerate: 
    pixels = physics.render()
    video.append(pixels.copy())
\end{lstlisting}
Plotting the movement trajectories, getting creature colours using \mono{bind()}:
\begin{lstlisting}[language=Python]
creature_colors = physics.bind(torsos).rgba[:, :3]
fig, ax = plt.subplots(figsize=(8, 8))
ax.set_prop_cycle(color=creature_colors)
ax.plot(pos_x, pos_y, linewidth=4)
\end{lstlisting}
\vspace{-.2cm}
\begin{figure}[H]
\floatbox[{\capbeside\thisfloatsetup{capbesideposition={left,top},
capbesidewidth=0.57\textwidth}}]{figure}[\FBwidth]
{\caption*{\normalsize
\href{https://youtu.be/0Lw_77PErjg}{\textsf{youtu.be/0Lw\_77PErjg}} shows a clip of the locomotion. The plot on the right shows the corresponding movement trajectories of creature positions. Note how \mono{physics.bind(torsos)} was used to access both \mono{xpos} and \mono{rgba} values. Once the \mono{Physics} had been instantiated by \mono{from\_mjcf\_model()}, the \mono{bind()} method will expose both the associated \mono{mjData} and \mono{mjModel} fields of an \mono{mjcf} element, providing unified access to all quantities in the simulation. 
}}
{\includegraphics[width=0.41\textwidth]{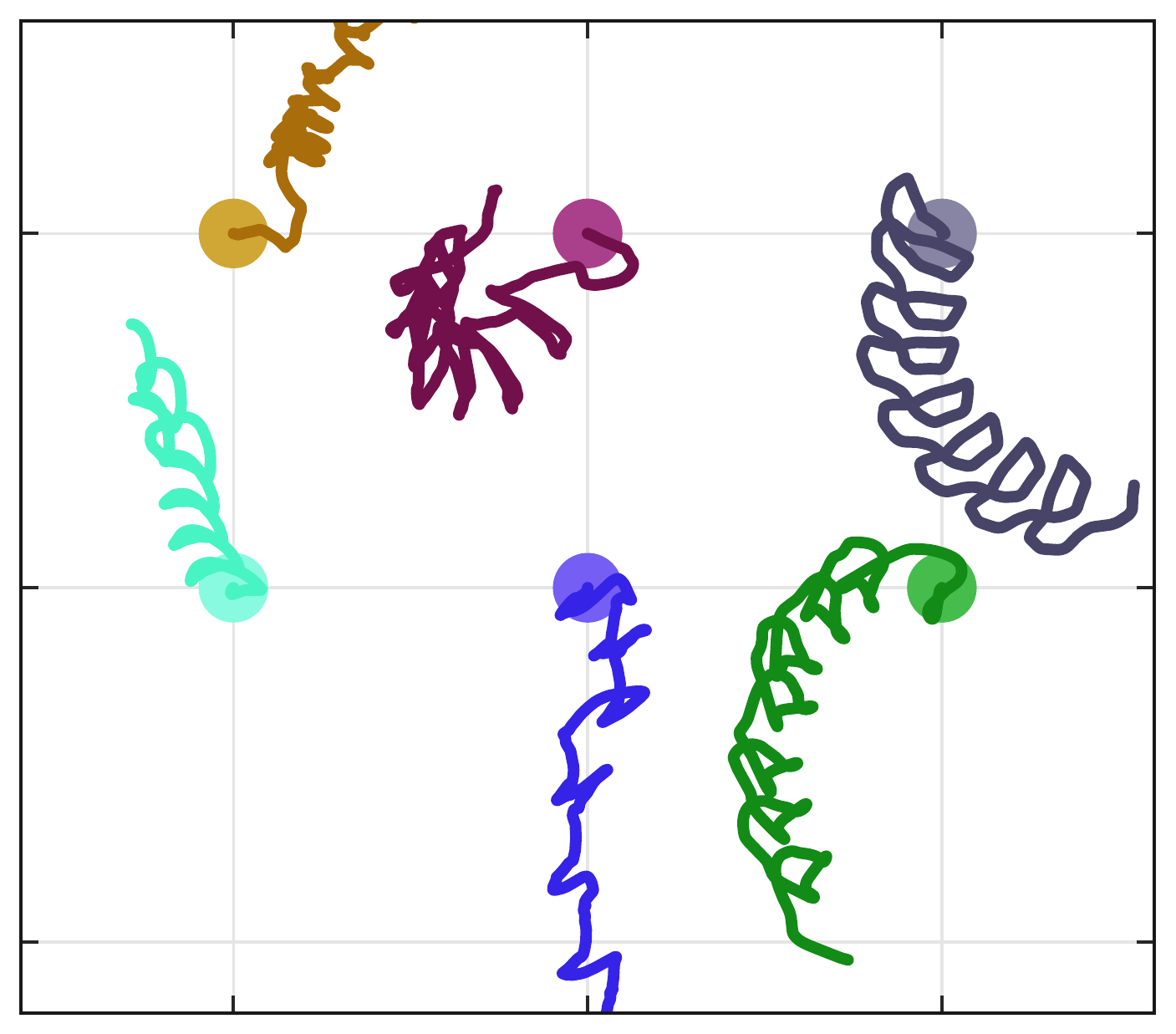}}
\end{figure}
\vspace{-.5cm}

\subsection{Debugging}

In order to aid in troubleshooting MJCF compilation problems on models that are assembled programmatically, PyMJCF implements a debug mode where individual XML elements and attributes can be traced back to the line of Python code that last modified it. This feature is not enabled by default since the tracking mechanism is expensive to run. However when a compilation error is encountered the user is instructed to restart the program with the \mono{--pymjcf\_debug} runtime flag. This flag causes PyMJCF to internally log the Python stack trace each time the model is modified. MuJoCo's error message is parsed to determine the line number in the generated XML document, which can be used to cross-reference to the XML element that is causing the error. The logged stack trace then allows PyMJCF to report the line of Python code that is likely to be responsible.

Occasionally, the XML compilation error arises from incompatibility between attached models or broken cross-references. Such errors are not necessarily local to the line of code that last modified a particular element. For such a scenario, PyMJCF provides an additional \mono{--pymjcf\_debug\_full\_dump\_dir} flag that causes the entirety of the internal stack trace logs to be written to files at the specified directory.

\section{Reinforcement learning interface}
\label{sec:rl}
Reinforcement learning is a computational framework wherein an \emph{agent}, through sequential interactions with an \emph{environment}, tries to learn a behaviour policy that maximises future rewards \citep{Sutton1998}.

 

\subsection{Reinforcement Learning API}

Environments in \dmcontrol adhere to DeepMind's \dmenv interface, defined in the  \href{https://github.com/deepmind/dm_env}{\textsf{github.com/deepmind/dm\_env}} repository. In brief, a run-loop using \dmenv may look like
\pagebreak
\begin{lstlisting}[language=Python]
for _ in range(num_episodes):
  timestep = env.reset()
  while True:
    action = agent.step(timestep)
    timestep = env.step(action)
    if timestep.last():
      agent.step(timestep)
      break
\end{lstlisting}
Each call to an environment's \mono{step()} method returns a \mono{TimeStep} namedtuple with \mono{step\_type}, \mono{reward}, \mono{discount} and \mono{observation} fields.
Each episode starts with a \mono{step\_type} of \mono{FIRST}, ends with a \mono{step\_type} of \mono{LAST}, and has a \mono{step\_type} of \mono{MID} for all intermediate timesteps. A \mono{TimeStep} also has corresponding \mono{first()}, \mono{mid()} and \mono{last()} methods, as illustrated above. Please see the \dmenv  \href{https://github.com/deepmind/dm_env/blob/master/docs/index.md}{repository documentation} for more details.

\subsubsection{The \texttt{Environment} class}
The class \mono{Environment}, found within the \mono{dm\_control.rl.control} module, implements the \dmenv environment interface:

\begin{description}[leftmargin=0cm,itemindent=0.2cm,labelsep=0.2cm, parsep=0cm]
\item[\mono{reset()}] Initialises the state, sampling from some initial state distribution.
\item[\mono{step()}] Accepts an action, advances the simulation by one time-step, and returns a \mono{TimeStep} namedtuple.
\item[\mono{action\_spec()}] describes the actions accepted by an \mono{Environment}.
The method returns an \mono{ArraySpec}, with attributes that describe the shape, data type, and optional lower and upper bounds for the action arrays. For example, random agent interaction can be implemented as
\begin{lstlisting}[language=Python]
spec = env.action_spec()
time_step = env.reset()
while not time_step.last():
  action = np.random.uniform(spec.minimum, spec.maximum, spec.shape)
  time_step = env.step(action)
\end{lstlisting}

\item[\mono{observation\_spec()}] returns an \mono{OrderedDict} of \mono{ArraySpec}s describing the shape and data type of each corresponding observation.
\end{description}
A \textbf{\mono{TimeStep}} namedtuple contains:
\begin{itemize}[parsep=0cm]
\item \mono{step\_type}, an enum with a value in \mono{[FIRST, MID, LAST]}.
\item \mono{reward}, a floating point scalar, representing the reward from the previous transition.
\item \mono{discount}, a scalar floating point number $\gamma \in [0, 1]$.
\item \mono{observation}, an \mono{OrderedDict} of \numpy arrays matching the specification returned by \mono{observation\_spec()}.
\end{itemize}
Whereas the \mono{step\_type} specifies whether or not the episode is terminating, it is the \mono{discount} $\gamma$ that determines the termination type. $\gamma=0$ corresponds to a terminal state\footnote{i.e.\ the sum of future reward is equal to the current reward.} as in the first-exit or finite-horizon formulations. A terminal \mono{TimeStep} with $\gamma=1$ corresponds to the infinite-horizon formulation; in this case an agent interacting with the environment should treat the episode as if it could have continued indefinitely, even though the sequence of observations and rewards is truncated. In this case a parametric value function may be used to estimate future returns.

\subsection{Reward functions}
\begin{figure}[h]
\includegraphics[width=0.98\textwidth]{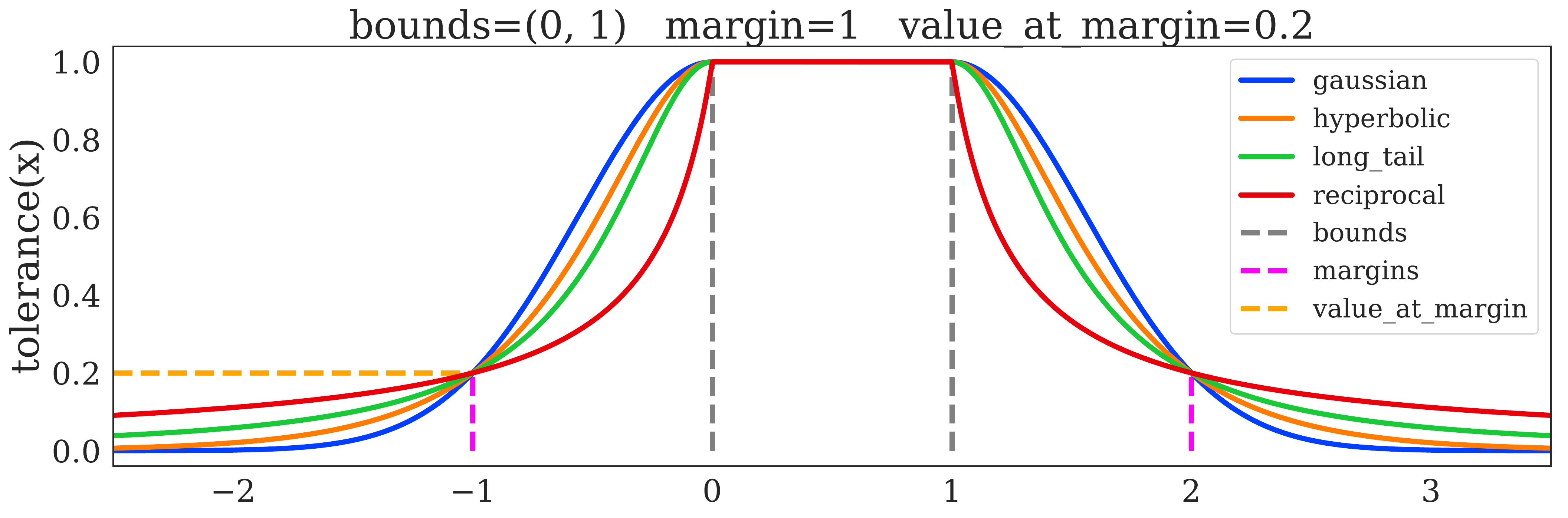}

\vspace{.5cm}
\includegraphics[width=0.98\textwidth]{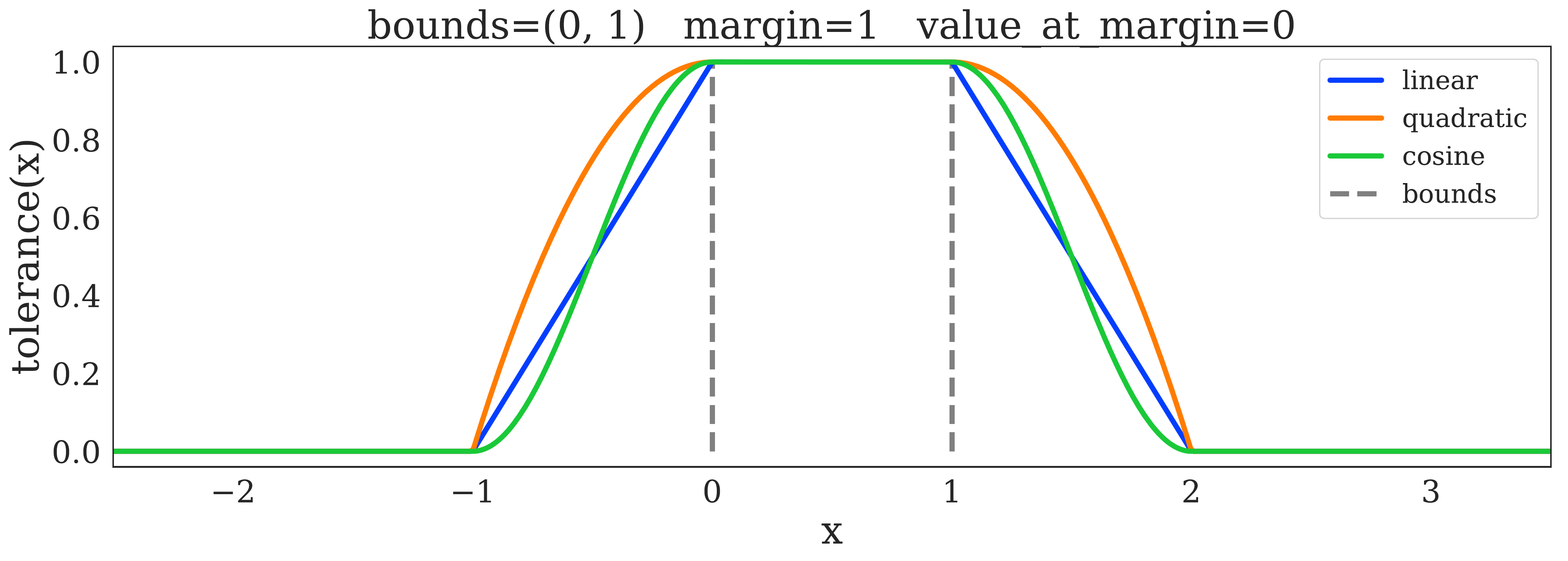}
\caption{The \mono{tolerance(x, bounds=(lower, upper))} function returns 1 if \mono{x} is within the \mono{bounds} interval and 0 otherwise. If the optional \mono{margin} argument is given, the output will decrease smoothly with distance from the interval, taking a value of \mono{value\_at\_margin} at a distance of \mono{margin}. Several types of sigmoid-like functions are available by setting the \mono{sigmoid} argument. \emph{Top:} Four infinite-support reward functions, for which \mono{value\_at\_margin} must be positive. \emph{Bottom:} Three finite-support reward functions with \mono{value\_at\_margin=0}.}
\label{fig:tolerance}
\end{figure}

Rewards in \dmcontrol tasks are in the unit interval, $r(\mathbf{s}, \mathbf{a}) \in [0, 1]$. Some tasks have ``sparse'' rewards, i.e., $r(\mathbf{s}, \mathbf{a}) \in \{0,1\}$. This structure is facilitated by the \mono{tolerance()} function, see Figure \ref{fig:tolerance}. Since terms output by \mono{tolerance()} are in the unit interval, both \emph{averaging} and \emph{multiplication} operations maintain that property, facilitating reward design.

\vspace{-.5cm}
\section{The \textsf{Composer} task definition library}
\label{sec:composer}
The \textsf{Composer} framework organises RL environments into a common structure and endows scene elements with optional event handlers. At a high level, \textsf{Composer} defines three main abstractions for task design:
\begin{itemize}
    \item \mono{composer.Entity} represents a reusable self-contained building block that consists of an MJCF model, observables (Section \ref{sec:observable}), and possibly callbacks executed at specific stages of the environment's life time, as detailed in Section \ref{sec:callbacks}. A collection of entities can be organised into a tree structure by attaching one or more child entities to a parent. The root entity is conventionally referred to as an ``arena'', and provides a fixed \mono{<worldbody>} for the final, combined MJCF model.
    \item \mono{composer.Task} consists of a tree of \mono{composer.Entity} objects that occupy the physical scene and provides reward, observation and termination methods. A task may also define callbacks to implement ``game logic'', e.g.\ to modify the scene in response to various events, and provide additional task observables.
    \item \mono{composer.Environment} wraps a \mono{composer.Task} instance with an RL environment that agents can interact with. It is responsible for compiling the MJCF model, triggering callbacks at appropriate points of an episode (see Section \ref{sec:callbacks}), and determining when to terminate, either through task-defined termination criteria or a user defined time limit. It also holds a random number generator state that is used by the callbacks, enabling reproducibility.
\end{itemize}
Section \ref{sec:observable} describes \mono{observable}, a \textsf{Composer} module  for exposing observations, supporting noise, buffering and delays. Section \ref{sec:variation} describes \mono{variation}, a module for implementing model variations. Section \ref{sec:callbacks} describes the callbacks used by \textsf{Composer} to implement these and additional user-defined behaviours. A self-contained \textsf{Composer} tutorial follows in Section~\ref{sec:composertutorial}

\subsection{The \texorpdfstring{\texttt{observable}}{observable} module}
\label{sec:observable}
An ``observable'' represents a quantity derived from the state of the simulation, that may be returned as an observation to the agent.
Observables may be bound to a particular entity (e.g.\ sensors belonging to a robot), or they may be defined at the task level.
The latter is often used for providing observations that relate to more than one entity in the scene (e.g.\ the distance between an end effector of a robot entity and a site on a different entity).
A particular entity may define any number of observables (such as joint angles, pressure sensors, cameras), and it is up to the task designer to select which of these should appear in the agent's observations.
Observables can be configured in a number of ways:

\begin{itemize}
    \item \mono{enabled}: (boolean)
    Whether the observable is computed and returned to the agent. Set to \mono{False} by default.

    \item \mono{update\_interval}: (integer or callable returning an integer)
    Specifies the interval, in simulation steps, at which the values of the observable will be updated.
    The last value will be repeated between updates.
    This parameter may be used to simulate sensors with different sample rates.
    Sensors with stochastic rates may be modelled by passing a callable that returns a random integer.

    \item \mono{buffer\_size}: (integer)
    Controls the size of the internal FIFO buffer used to store observations that were sampled on previous simulation time-steps.
    In the default case where no \mono{aggregator} is provided (see below), the entire contents of the buffer is returned as an observation at each control timestep.
    This can be used to avoid discarding observations from sensors whose values may change significantly within the control timestep.
    If the buffer size is sufficiently large, it will contain observations from  previous control timesteps, endowing the environment with a simple form of memory.

    \item \mono{corruptor}: (callable)
    Performs a point-wise transformation of each observation value before it is inserted into the buffer.
    Corruptors are most commonly used to simulate observation noise.

    \item \mono{aggregator}: (callable or predefined string)
    Performs a reduction over all of the elements in the observation buffer.
    For example this can be used to take a moving average over previous observation values.

    \item \mono{delay}: (integer or callable returning an integer)
    Specifies a delay (in terms of simulation timesteps) between when the value of the observable is sampled, and when it will appear in the observations returned by the environment.
    This parameter can be used to model sensor latency.
    Stochastic latencies may be modelled by passing a callable that returns a randomly sampled integer.
\end{itemize}
During each control step the evaluation of observables is optimized such that only callables for observables that can appear in future observations are evaluated.
For example, if we have an observable with \mono{update\_interval=1} and \mono{buffer\_size=1} then it will only be evaluated once per control step, even if there are multiple simulation steps per control step.
This avoids the overhead of computing intermediate observations that would be discarded.


\subsection{The \texorpdfstring{\texttt{variation}}{variation} module}
\label{sec:variation}
To improve the realism of the simulation, it is often desirable to randomise elements of the environment, especially those whose values are uncertain.
Stochasticity can be added to both the observables e.g.\ sensor noise (the \mono{corruptor} of the previous section), as well as the model itself (a.k.a.\ ``domain randomisation''). The latter is a popular method for increasing the robustness learned control policies.
The \mono{variation} module provides methods to add and configure forms of stochasticity in the \textsf{Composer} framework.
The following base API is provided:
\begin{itemize}
    \item \mono{Variation}:
    The base class.
    Subclasses should implement the abstract method \mono{\_\_call\_\_(self, initial\_value, current\_value, random\_state)},\newline which returns a numerical value, possibly depending on an \mono{initial\_value} (e.g. original geom mass) and \mono{current\_value} (e.g. previously sampled geom mass).
    \mono{Variation} objects support arithmetic operations with numerical primitives and other \mono{Variation} objects.
    \item \mono{MJCFVariator}:
    A class for varying attributes of MJCF elements, e.g.\ geom size.
    The \mono{MJCFVariator} keeps track of initial and current attribute values and passes them to the \mono{Variation} object.
    It should be called in the \mono{initialize\_episode\_mjcf} stage, before the model is compiled.
    \item \mono{PhysicsVariator}:
    Similar to \mono{MJCFVariator}, except for bound attributes, e.g. external forces.
    Should be called in the \mono{initialize\_episode} stage after the model has been compiled.
    \item \mono{evaluate}:
    Method to traverse an arbitrarily nested structure of callables or constant values, and evaluate any callables (such as \mono{Variation} objects).    
\end{itemize}
A number of submodules provide classes for commonly occurring use cases:
\begin{itemize}
    \item \mono{colors}:
    Used to define variations in different colour spaces, such as RGB, HSV and grayscale.
    \item \mono{deterministic}:
    Deterministic variations such as constant and fixed sequences of values, in case more control over the exact values is required.
    \item \mono{distributions}:
    Wraps a number of distributions available in \mono{numpy.random} as \mono{Variation} objects.
    Any distribution parameters passed can themselves also be \mono{Variation} objects.
    \item \mono{noises}:
    Used to define additive and multiplicative noise using the distributions mentioned above, e.g. for modelling sensor noise.
    \item \mono{rotations}:
    Useful for defining variations in quaternion space, e.g. random rotation on a \mono{composer.Entity}'s pose.
\end{itemize}

\subsection{The \textsf{Composer} callback lifecycle}
\label{sec:callbacks}
\tikzstyle{event2} = [draw, rectangle,fill=blue!10, align=center]
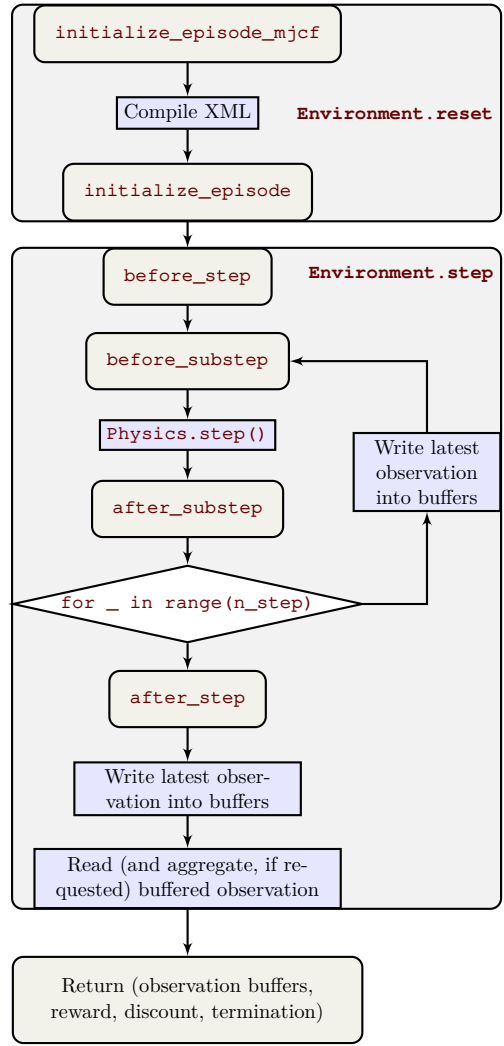
\begin{wrapfigure}[35]{r}[-1.5cm]{5cm}
\vspace{-.3cm}
\ffigbox[0.9\FBwidth]
{
\caption{
Diagram showing the life-cycle of \textsf{Composer} callbacks. Rounded rectangles represent callbacks that Tasks and Entities may implement. Blue rectangles represent built-in \textsf{Composer} operations.}
}
{
\label{fig:callbacks}
{\begin{tikzpicture}[node distance = 1.6cm, auto, thick,scale=1, every node/.style={scale=.7}]
    \node [hook] (initialize_episode_mjcf) {\mono{initialize\_episode\_mjcf}};
    \node [event2, below of=initialize_episode_mjcf,yshift=0.1cm] (compile) {Compile XML};
    \node [hook, below of=compile,yshift=0.1cm] (initialize_episode) {\mono{initialize\_episode}};
    \node [hook, below of=initialize_episode] (before_step) {\mono{before\_step}};
    \node [hook, below of=before_step] (before_substep) {\mono{before\_substep}};
    \node [event2, below of=before_substep,yshift=0.2cm] (mj_step) {\mono{Physics.step()}};
    \node [hook, below of=mj_step,yshift=0.2cm] (after_substep) {\mono{after\_substep}};

    \node [decision, below of=after_substep, yshift=-0.2cm] (enough_steps) {\mono{for \_ in range(n\_step)}};
    \node [event2, right of=mj_step, node distance=4.5cm, text width=2.5cm, yshift=-0.7cm] (update_obs_substep) {Write latest\\observation\\into buffers};
    \node [hook, below of=enough_steps, yshift=-0.2cm] (after_step) {\mono{after\_step}};

    \node [event2, below of=after_step, text width=4cm, yshift=-0.1cm] (update_obs_step) {Write latest observation into buffers};
    \node [event2, below of=update_obs_step, text width=5.5cm, yshift=-0.1cm] (read_obs) {Read (and aggregate, \linebreak[1] if requested) buffered observation};
    \node [hook, below of=read_obs, text width=5.8cm, yshift=-.7cm] (return) {
    \vbox {
    Return (observation buffers,
        reward,
        discount,
        termination)
    }};

    \path [line] (initialize_episode_mjcf) -- (compile);
    \path [line] (compile) -- (initialize_episode);
    \path [line] (initialize_episode) -- (before_step);
    \path [line] (before_step) -- (before_substep);
    \path [line] (before_substep) -- (mj_step);
    \path [line] (mj_step) -- (after_substep);
    \path [line] (after_substep) -- (enough_steps);
    \path [line] (enough_steps) -| node [near start] {} (update_obs_substep);
    \path [line] (update_obs_substep) |- (before_substep);
    \path [line] (enough_steps) -- node {} (after_step);
    \path [line] (after_step) -- (update_obs_step);
    \path [line] (update_obs_step) -- (read_obs);
    \path [line] (read_obs) -- (return);

    \begin{pgfonlayer}{background}
        \filldraw [fill=black!5,draw=black,rounded corners,thick]
        (initialize_episode.south -| enough_steps.west) rectangle
        (initialize_episode_mjcf.north -| update_obs_substep.east)
        node[anchor=north east, yshift=-1.8cm](env_reset){\mono{\textbf{Environment.reset}}};
        
        \filldraw [fill=black!5,draw=black,rounded corners,thick]
        (read_obs.south -| enough_steps.west) rectangle
        (before_step.north -| update_obs_substep.east)
        node[anchor=north east,yshift=-.2cm](env_step){\mono{\textbf{Environment.step}}};
    \end{pgfonlayer}
\end{tikzpicture}}
}
\end{wrapfigure}

Figure~\ref{fig:callbacks} illustrates the lifecycle of \textsf{Composer} callbacks.
These can be organised into those that occur when an RL episode is reset, and those that occur when the environment is stepped. For a given callback, the one that is defined in the Task is executed first, followed by those defined in each \mono{Entity} following depth-first traversal of the Entity tree starting from the root (which by convention is the arena) and the order in which Entities were attached.

The first of the two callbacks in 
\mono{reset} is \mono{initialize\_episode\_mjcf}, which allows the MJCF model to be modified between episodes. It is useful for changing quantities that are fixed once the model has been compiled. These modifications affect the generated XML which is 
then compiled into a \mono{Physics} instance and passed to the \mono{initialize\_episode} callback, where the initial state can be set.

The \mono{Environment.step} sequence begins at the \mono{before\_step} callback. One key role of this callback is to translate agent actions into the \mono{Physics} control vector. 

To guarantee stability, it is often necessary to reduce the time-step of the physics simulation. In order to decouple these possibly very small steps and the agent's control time-step, we introduce a substep loop. Each \mono{Physics} substep is preceded by \mono{before\_substep} and followed by \mono{after\_substep}. These callbacks are useful for detecting transient events that may occur in the middle of an environment step, e.g. a button press. The internal observation buffers are then updated according to the configured \mono{update\_interval} of each individual \mono{Observable}, unless the substep happens to be the last one in the environment step, in which case the \mono{after\_step} callback is called first before the final update of the observation buffers. The internal observation buffers are then processed according to the \mono{delay}, \mono{buffer\_size}, and \mono{aggregator} settings of each \mono{Observable} to generate ``output buffers'' that are returned externally.

At the end of each \mono{Environment.step}, the Task's \mono{get\_reward}, \mono{get\_discount}, and \mono{should\_terminate\_episode} callbacks are called in order to obtain the step's reward, discount, and termination status respectively. Usually, the these three are not entirely independent of each other, and it is therefore recommended to compute all of these in the \mono{after\_step} callback, cache the values in the Task instance, and return them in the respective callbacks.

\subsection{\textsf{Composer} tutorial}
\label{sec:composertutorial}
In this tutorial we will create a task requiring our ``creature'' from Section \ref{sec:pymjcftutorial} to press a colour-changing button on the floor with a prescribed force. We begin by implementing our ``creature'' as a \mono{composer.Entity}:

\begin{lstlisting}[language=Python]
from dm_control import composer
from dm_control.composer.observation import observable

class Creature(composer.Entity):
  """A multi-legged creature derived from `composer.Entity`."""
  def _build(self, num_legs):
    self._model = make_creature(num_legs)

  def _build_observables(self):
    return CreatureObservables(self)

  @property
  def mjcf_model(self):
    return self._model
      
  @property
  def actuators(self):
    return tuple(self._model.find_all('actuator'))

class CreatureObservables(composer.Observables):
  """Add simple observable features for joint angles and velocities."""
  @composer.observable
  def joint_positions(self):
    all_joints = self._entity.mjcf_model.find_all('joint')
    return observable.MJCFFeature('qpos', all_joints)

  @composer.observable
  def joint_velocities(self):
    all_joints = self._entity.mjcf_model.find_all('joint')
    return observable.MJCFFeature('qvel', all_joints)
\end{lstlisting}
The \mono{Creature} Entity includes generic Observables for joint angles and velocities. Because \mono{find\_all()} is called on the \mono{Creature}'s MJCF model, it will only return the creature's leg joints, and not the ``free'' joint with which it will be attached to the world. Note that \textsf{Composer} Entities should override the \mono{\_build} and \mono{\_build\_observables} methods rather than \mono{\_\_init\_\_}. The implementation of \mono{\_\_init\_\_} in the base class calls \mono{\_build} and \mono{\_build\_observables}, in that order, to ensure that the entity's MJCF model is created before its observables. This was a design choice which allows the user to refer to an observable as an attribute (\mono{entity.observables.foo}) while still making it clear which attributes are observables. The stateful \mono{Button} class derives from \mono{composer.Entity} and implements the \mono{initialize\_episode} and \mono{after\_substep} callbacks.
\begin{lstlisting}[language=Python]
NUM_SUBSTEPS = 25  # The number of physics substeps per control timestep.
class Button(composer.Entity):
  """A button Entity which changes colour when pressed with certain force."""
  def _build(self, target_force_range=(5, 10)):
    self._min_force, self._max_force = target_force_range
    self._mjcf_model = mjcf.RootElement()
    self._geom = self._mjcf_model.worldbody.add(
        'geom', type='cylinder', size=[0.25, 0.02], rgba=[1, 0, 0, 1])
    self._site = self._mjcf_model.worldbody.add(
        'site', type='cylinder', size=self._geom.size*1.01, rgba=[1, 0, 0, 0])
    self._sensor = self._mjcf_model.sensor.add('touch', site=self._site)
    self._num_activated_steps = 0

  def _build_observables(self):
    return ButtonObservables(self)

  @property
  def mjcf_model(self):
    return self._mjcf_model
  def _update_activation(self, physics):
    """Update the activation and colour if the desired force is applied."""
    current_force = physics.bind(self.touch_sensor).sensordata[0]
    is_activated = (current_force >= self._min_force and
                    current_force <= self._max_force)
    red = [1, 0, 0, 1]
    green = [0, 1, 0, 1]
    physics.bind(self._geom).rgba = green if is_activated else red
    self._num_activated_steps += int(is_activated)

  def initialize_episode(self, physics, random_state):
    self._reward = 0.0
    self._num_activated_steps = 0
    self._update_activation(physics)

  def after_substep(self, physics, random_state):
    self._update_activation(physics)

  @property
  def touch_sensor(self):
    return self._sensor

  @property
  def num_activated_steps(self):
    return self._num_activated_steps

class ButtonObservables(composer.Observables):
  """A touch sensor which averages contact force over physics substeps."""
  @composer.observable
  def touch_force(self):
    return observable.MJCFFeature('sensordata', self._entity.touch_sensor,
                                  buffer_size=NUM_SUBSTEPS, aggregator='mean')
\end{lstlisting}
Note how the Button counts the number of sub-steps during which it is pressed with the desired force. It also exposes an \mono{Observable} of the force being applied to the button, whose value is an average of the readings over the physics time-steps.
\newpage
\noindent We import some \mono{variation} modules and an arena factory (see Section \ref{sec:locomotion}):
\begin{lstlisting}[language=Python]
from dm_control.composer import variation
from dm_control.composer.variation import distributions
from dm_control.composer.variation import noises
from dm_control.locomotion.arenas import floors
\end{lstlisting}
\noindent A simple \mono{Variation} samples the initial position of the target:
\begin{lstlisting}[language=Python]
class UniformCircle(variation.Variation):
  """ A uniformly sampled horizontal point on a circle of radius `distance`"""
  def __init__(self, distance):
    self._distance = distance
    self._heading = distributions.Uniform(0, 2*np.pi)

  def __call__(self, initial_value=None, 
               current_value=None, random_state=None):
    distance, heading = variation.evaluate(
        (self._distance, self._heading), random_state=random_state)
    return (distance*np.cos(heading), distance*np.sin(heading), 0)
\end{lstlisting}
\noindent We will now define the \mono{PressWithSpecificForce} Task, which combines all the above elements. The \mono{\_\_init\_\_} constructor sets up the scene:
\begin{lstlisting}[language=Python]
class PressWithSpecificForce(composer.Task):

  def __init__(self, creature):
    self._creature = creature
    self._arena = floors.Floor()
    self._arena.add_free_entity(self._creature)
    self._arena.mjcf_model.worldbody.add('light', pos=(0, 0, 4))
    self._button = Button()
    self._arena.attach(self._button)

    # Configure initial poses
    self._creature_initial_pose = (0, 0, 0.15)
    button_distance = distributions.Uniform(0.5, .75)
    self._button_initial_pose = UniformCircle(button_distance)

    # Configure variators
    self._mjcf_variator = variation.MJCFVariator()
    self._physics_variator = variation.PhysicsVariator()

    # Configure and enable observables
    pos_corruptor = noises.Additive(distributions.Normal(scale=0.01))
    self._creature.observables.joint_positions.corruptor = pos_corruptor
    self._creature.observables.joint_positions.enabled = True
    vel_corruptor = noises.Multiplicative(distributions.LogNormal(sigma=0.01))
    self._creature.observables.joint_velocities.corruptor = vel_corruptor
    self._creature.observables.joint_velocities.enabled = True
    self._button.observables.touch_force.enabled = True

    # Add button position observable in the Creature's egocentric frame
    self._task_observables = {}
    def to_button(physics):
      button_pos, _ = self._button.get_pose(physics)
      return self._creature.global_vector_to_local_frame(physics, button_pos)
    self._task_observables['button_position'] = observable.Generic(to_button)
    for obs in self._task_observables.values():
      obs.enabled = True  # Enable all observables.

    self.control_timestep = NUM_SUBSTEPS * self.physics_timestep
# Continued below...
\end{lstlisting}

\newpage
\noindent Continuing the \mono{PressWithSpecificForce} Task definition, we now implement our \textsf{Composer} callbacks, including the reward function:
\begin{lstlisting}[language=Python]
# Continued from above...
  @property
  def root_entity(self):
    return self._arena

  @property
  def task_observables(self):
    return self._task_observables

  def initialize_episode_mjcf(self, random_state):
    self._mjcf_variator.apply_variations(random_state)

  def initialize_episode(self, physics, random_state):
    self._physics_variator.apply_variations(physics, random_state)
    creature_pose, button_pose = variation.evaluate(
        (self._creature_initial_pose, self._button_initial_pose),
        random_state=random_state)
    self._creature.set_pose(physics, position=creature_pose)
    self._button.set_pose(physics, position=button_pose)

  def get_reward(self, physics):
    return self._button.num_activated_steps / NUM_SUBSTEPS
\end{lstlisting}
Finally, we can instantiate a \mono{Creature} Entity,
\begin{lstlisting}[language=Python]
creature = Creature(num_legs=4)
\end{lstlisting}
pass it to the \mono{PressWithSpecificForce} constructor to instantiate the task,
\begin{lstlisting}[language=Python]
task = PressWithSpecificForce(creature)
\end{lstlisting}
and expose it as an environment complying with the \mono{dm\_env.Environment} API:
\begin{lstlisting}[language=Python]
env = composer.Environment(task)
\end{lstlisting}
\vspace{-.3cm}
\begin{figure}[H]
\floatbox[{\capbeside\thisfloatsetup{capbesideposition={left,top},
capbesidewidth=0.57\textwidth}}]{figure}[\FBwidth]
{\caption*{\normalsize
Here is our creature with a large red button, waiting to be pressed.
}}
{\includegraphics[width=0.4\textwidth]{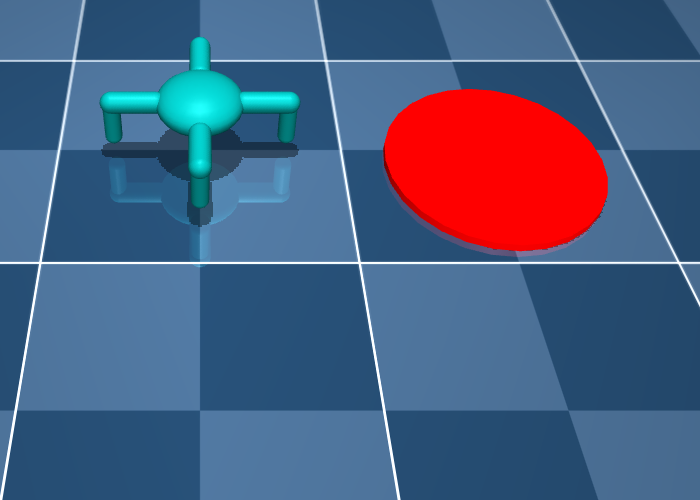}}
\end{figure}

\pagebreak
\part{Tasks}

\section{The \textsf{Control Suite}}
\label{sec:controlsuite}
The \textsf{Control Suite} is a set of stable, well-tested tasks designed to serve as a benchmark for continuous control learning agents. Tasks are written using the basic \textsf{MuJoCo} interface of Section~\ref{sec:mujoco}. Standardised action, observation and reward structures make suite-wide benchmarking simple and learning curves easy to interpret. Unlike the more elaborate domains of the Sections \ref{sec:locomotion} and \ref{sec:manipulation}, \textsf{Control Suite} domains are not meant to be modified, in order to facilitate benchmarking. For more details regarding benchmarking, refer to our original publication \citep{deepmindcontrolsuite2018}. A video montage of \textsf{Control Suite} domains can be found at \href{https://www.youtube.com/watch?v=rAai4QzcYbs}{\textsf{youtu.be/rAai4QzcYbs}}.

\subsection{\textsf{Control Suite} design conventions}

\begin{description}[leftmargin=0cm]

\item[Action:]
With the exception of the LQR domain (see below), the action vector is in the unit box, i.e., $\mathbf{a}\in \mathcal{A} \equiv \left[-1, 1\right]^{\dim(\mathcal{A})}$.

\item[Dynamics:]
While the state notionally evolves according to a continuous ordinary differential equation $\dot{\mathbf{s}} = \mathbf{f}_c(\mathbf{s},\mathbf{a})$, in practice temporal integration is discrete\footnote{Most domains use \textsf{MuJoCo}'s default semi-implicit Euler integrator. A few domains which have smooth dynamics use 4th-order Runge Kutta.} with some fixed, finite time-step: $\mathbf{s}_{t+h} = \mathbf{f}(\mathbf{s}_t,\mathbf{a}_t)$.

\item[Observation:]
When using the default observations (rather than pixels, see below), all tasks\footnote{With the exception of \mono{point-mass:hard} (see below).} are strongly observable, i.e. the state can be recovered from a single observation. Observation features which depend only on the state (position and velocity) are functions of the current state. Features which are also dependent on controls (e.g.\ touch sensor readings) are functions of the previous transition. 

\item[Reward:]
Rewards in the Control Suite, with the exception of the LQR domain, are in the unit interval, i.e., $r(\mathbf{s}, \mathbf{a}) \in [0, 1]$. Some rewards are ``sparse'', i.e., $r(\mathbf{s}, \mathbf{a}) \in \{0,1\}$. This structure is facilitated by the \mono{tolerance()} function, see Figure~\ref{fig:tolerance}. 

\item[Termination and Discount:]
Control problems are usually classified as finite-horizon, first-exit and infinite-horizon~\citep{bertsekas1995dynamic}. Control Suite tasks have no terminal states or time limit and are therefore of the infinite-horizon variety. Notionally the objective is the infinite-horizon average return $\lim_{T\rightarrow \infty} T^{-1}\int_{0}^{T}r(\mathbf{s}_t,\mathbf{a}_t)dt$, but in practice our agents internally use the discounted formulation $\int_{0}^{\infty}e^{-t/\tau}r(\mathbf{s}_t,\mathbf{a}_t)dt$ or in discrete time
$\sum_{i=0}^\infty \gamma^i r(\mathbf{s}_i,\mathbf{a}_i)$,
where $\gamma=e^{-h/\tau}$ is the discount factor. In the limit $\tau \rightarrow \infty$ (equivalently $\gamma \rightarrow 1$), the policies of the discounted-horizon and average-return formulations are identical. All Control Suite tasks with the exception of \mono{LQR}\footnote{The \mono{LQR} task terminates with $\gamma=0$ when the state is very close to the origin, as a proxy for the exponential convergence of stabilised linear systems.} return $\gamma=1$ at every step, including on termination.

\item[Evaluation:]
While agents are expected to optimise for infinite-horizon returns, these are difficult to measure. As a proxy we use fixed-length episodes of 1000 time steps. Since all reward functions are designed so that $r \approx 1$ near goal states, learning curves measuring total returns can all have the same y-axis limits of $[0, 1000]$, making them easier to interpret and to average over all tasks. While a perfect score of 1000 is not usually achievable, scores outside the $[800, 1000]$ range can be confidently said to be sub-optimal.
\end{description}

\subsubsection*{Model and Task verification}

Verification in this context means making sure that the physics simulation is stable and that the task is solvable: 

\begin{itemize}
\item Simulated physics can easily destabilise and diverge, mostly due to errors introduced by time discretisation. Smaller time-steps are more stable, but require more computation per unit of simulation time, so the choice of time-step is always a trade-off between stability and speed~\citep{erez2015simulation}. What's more, learning agents are  very good at discovering and exploiting instabilities.\footnote{This phenomenon, known as \textit{Sims' Law}, was first articulated in \citep{sims1994evolving}: ``Any bugs that allow energy leaks from non-conservation, or even round-off errors, will inevitably be discovered and exploited''.}

\item It is surprisingly easy to write tasks that are much easier or harder than intended, that are impossible to solve or that can be solved by very different strategies than expected (i.e.\ ``cheats''). To prevent these situations, the Atari{\scriptsize ™} games that make up ALE were extensively tested over more than 10 man-years\footnote{Marc Bellemare, personal communication.}. However, many continuous control domains cannot be solved by humans with standard input devices, due to the large action space, so a different approach must be taken.
\end{itemize}

\noindent In order to tackle both of these challenges, we ran variety of learning agents (e.g.\ \citealt{lillicrap2015continuous, mnih2016asynchronous}) against all tasks, and iterated on each task's design until we were satisfied that the physics was stable and non-exploitable, and that the task is solved correctly by at least one agent. Tasks that were solvable were collated into the \mono{benchmarking} set. Tasks which were not yet solved at the time of development are in the \mono{extra} set of tasks.

\subsection*{The \mono{suite} module}

To load an environment representing a task from the suite, use \mono{suite.load()}:
\begin{lstlisting}[language=Python]
from dm_control import suite

# Load one task:
env = suite.load(domain_name="cartpole", task_name="swingup")

# Iterate over a task set:
for domain_name, task_name in suite.BENCHMARKING:
  env = suite.load(domain_name, task_name)
  ...
\end{lstlisting}
Wrappers can be used to modify the behaviour of environments:

\begin{description}[leftmargin=0cm,itemindent=0cm]

\item{\textbf{Pixel observations:}}
By default, Control Suite environments return feature observations. The \mono{pixel.Wrapper} adds or replaces these with images.
\begin{lstlisting}[language=Python]
from dm_control.suite.wrappers import pixels
env = suite.load("cartpole", "swingup")
# Replace existing features by pixel observations:
env_only_pixels = pixels.Wrapper(env)
# Pixel observations in addition to existing features.
env_plus_pixels = pixels.Wrapper(env, pixels_only=False)
\end{lstlisting}

\item{\textbf{Reward visualisation:}}
Models in the Control Suite use a common set of colours and textures for visual uniformity. As illustrated in the \href{https://youtu.be/rAai4QzcYbs}{video}, this also allows us to modify colours in proportion to the reward, providing a convenient visual cue. 
\begin{lstlisting}[language=Python]
env = suite.load("fish", "swim", task_kwargs, visualize_reward=True)
\end{lstlisting}

\end{description}

\subsection{Domains and Tasks} \label{sec:csdomains}
A \textbf{domain} refers to a physical model, while a \textbf{task} refers to an instance of that model with a particular MDP structure. For example the difference between the \mono{swingup} and \mono{balance} tasks of the \mono{cartpole} domain is whether the pole is initialised pointing downwards or upwards, respectively. In some cases, e.g.\ when the model is procedurally generated, different tasks might have different physical properties. Tasks in the Control Suite are collated into tuples according to predefined tags. Tasks used for benchmarking are in the \mono{BENCHMARKING} tuple (Figure~\ref{fig:benchmarking}), while those not used for benchmarking (because they are particularly difficult, or because they do not conform to the standard structure) are in the \mono{EXTRA} tuple. All suite tasks are accessible via the \mono{ALL\_TASKS} tuple. In the domain descriptions below, names are followed by three integers specifying the dimensions of the \textbf{state}, \textbf{control} and \textbf{observation} spaces: $\textrm{\bf Name } \Bigl(\dim(\mathcal{S}),\dim(\mathcal{A}),\dim(\mathcal{O})\Bigr)$. 

\vspace{-.2cm}
\myfigure{
\textbf{Pendulum (2, 1, 3):} The classic inverted pendulum. The torque-limited actuator is $1/6^{\mathrm{th}}$ as strong as required to lift the mass from motionless horizontal, necessitating several swings to swing up and balance. The \mono{swingup} task has a simple sparse reward: 1 when the pole is within $30^\circ$ of the vertical position and 0 otherwise.
}{./figures/pendulum.png}

\myfigure{
\textbf{Acrobot (4, 1, 6):} The underactuated double pendulum, torque applied to the second joint. The goal is to swing up and balance. Despite being low-dimensional, this is not an easy control problem. The physical model conforms to \citep{coulom2002reinforcement} rather than the earlier \citealt{spong1995swing}. The \mono{swingup} and \mono{swingup\_sparse} tasks have smooth and sparse rewards, respectively.
}{./figures/acrobot.png}

\myfigure{
\textbf{Cart-pole (4, 1, 5):} Swing up and balance an unactuated pole by applying forces to a cart at its base. The physical model conforms to \citealt{barto1983neuronlike}. Four benchmarking tasks: in \mono{swingup} and \mono{swingup\_sparse} the pole starts pointing down while in \mono{balance} and \mono{balance\_sparse} the pole starts near the upright position. }{./figures/cart-pole.png}

\begin{figure}[H]
\floatbox[{\capbeside\thisfloatsetup{capbesideposition={right,top},
capbesidewidth=0.585\textwidth}}]{figure}[\FBwidth]
{\caption*{
\textbf{Cart-k-pole ($\mathbf{2k\!+\!2,\; 1,\; 3k\!+\!2}$):}
The cart-pole domain allows to procedurally adding more poles, connected serially.
Two non-benchmarking tasks, \mono{two\_poles} and \mono{three\_poles} are available.
}}
{\includegraphics[width=2.5cm]{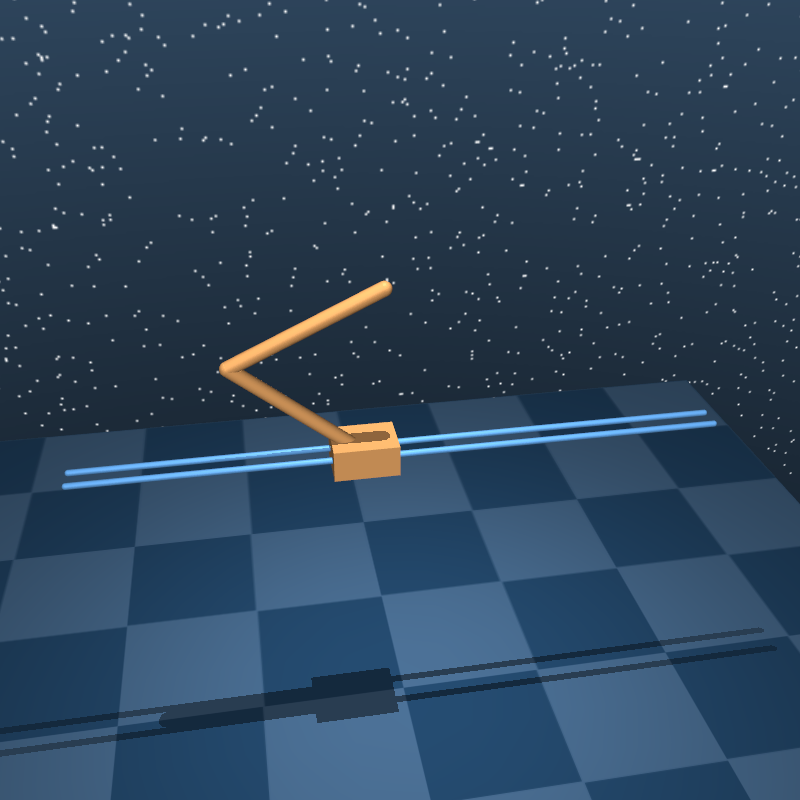}
\hspace{.1cm}
\includegraphics[width=2.5cm]{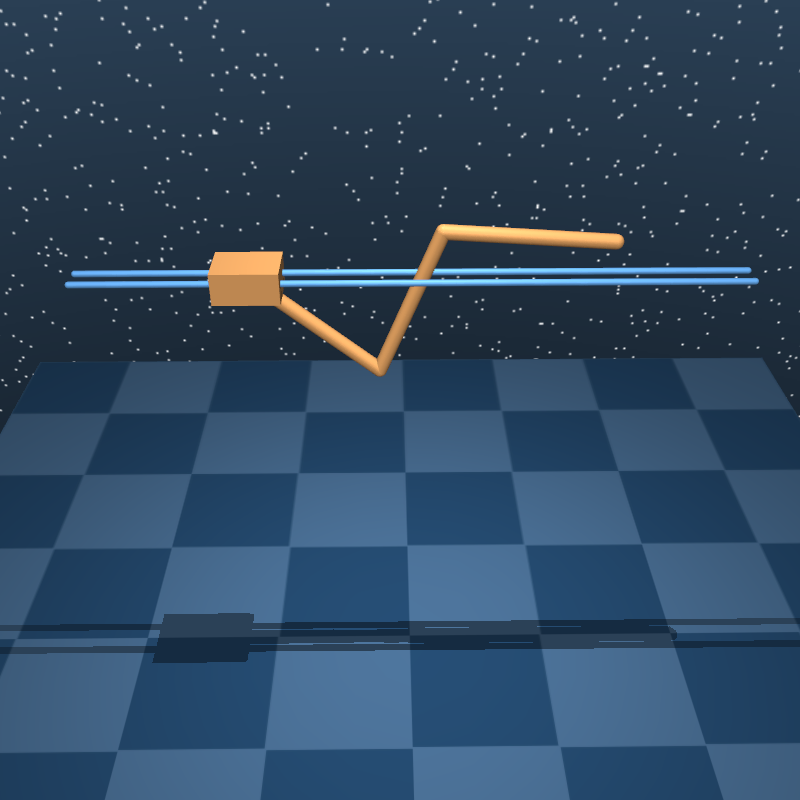}}
\end{figure}
\vspace{-.5cm}

\myfigure{
\textbf{Ball in cup (8, 2, 8):} A planar ball-in-cup task. An actuated planar receptacle can translate in the vertical plane in order to swing and catch a ball attached to its bottom. The \mono{catch} task has a sparse reward: 1 when the ball is in the cup, 0 otherwise. 
}{./figures/ball-in-cup.png}

\myfigure{
\textbf{Point-mass (4, 2, 4):} A planar point mass receives a reward of 1 when within a target at the origin. In the \mono{easy} task, one of simplest in the suite, the two actuators correspond to the global $x$ and $y$ axes. In the \mono{hard} task, the gain matrix from the controls to the axes is randomised for each episode, making it impossible to solve by memoryless agents. This task is not in the \mono{benchmarking} set. 
}{./figures/point-mass.png}

\myfigure{
\textbf{Reacher (4, 2, 6):} The simple two-link planar reacher with a randomised target location. The reward is one when the end effector penetrates the target sphere. In the \mono{easy} task the target sphere is bigger than on the \mono{hard} task (shown on the left).  
}{./figures/reacher.png}

\myfigure{
\textbf{Finger (6, 2, 12):} 
A 3-DoF toy manipulation problem based on \citep{tassa2010stochastic}. A planar `finger' is required to rotate a body on an unactuated hinge. In the \mono{turn\_easy} and \mono{turn\_hard} tasks, the tip of the free body must overlap with a target (the target is smaller for the \mono{turn\_hard} task). In the \mono{spin} task, the body must be continually rotated.
}{./figures/finger.png}

\myfigure{
\textbf{Hopper (14, 4, 15):} The planar one-legged hopper introduced in \citep{lillicrap2015continuous}, initialised in a random configuration. In the \mono{stand} task it is rewarded for bringing its torso to a minimal height. In the \mono{hop} task it is rewarded for torso height and forward velocity. 
}{./figures/hopper.png}

\myfigure{
\textbf{Fish (26, 5, 24):} A fish is required to swim to a target. This domain relies on \textsf{MuJoCo}'s simplified fluid dynamics. There are two tasks: in the \mono{upright} task, the fish is rewarded only for righting itself with respect to the vertical, while in the \mono{swim} task it is also rewarded for swimming to the target.
}{./figures/fish.png}

\myfigure{
\textbf{Cheetah (18, 6, 17):} A running planar biped based on \citep{wawrzynski2009real}.
The reward $r$ is linearly proportional to the forward velocity $v$ up to a maximum of $10 \mathrm{m/s}$ i.e.\ $r(v) = \max\bigl(0, \min(v/10, 1)\bigr)$.
}{./figures/cheetah.png}

\myfigure{
\textbf{Walker (18, 6, 24):} An improved planar walker based on the one introduced in \citep{lillicrap2015continuous}. In the \mono{stand} task reward is a combination of terms encouraging an upright torso and some minimal torso height. The \mono{walk} and \mono{run} tasks include a component encouraging forward velocity.
}{./figures/walker.png}

\myfigure{
\textbf{Manipulator (22, 5, 37):} A planar manipulator is rewarded for bringing an object to a target location. In order to assist with exploration, in 10\% of episodes the object is initialised in the gripper or at the target. Four \mono{manipulator} tasks: \mono{\{bring,insert\}\_\{ball,peg\}} of which only \mono{bring\_ball} is in the \mono{benchmarking} set. The other three are shown below.

}{./figures/manipulator.png}

\begin{figure}[H]
\floatbox[{\capbeside\thisfloatsetup{capbesideposition={right,top},
capbesidewidth=0.415\textwidth}}]{figure}[\FBwidth]
{\caption*{
\textbf{Manipulator extra:}
\mono{insert\_ball}: place the ball inside the basket. \mono{bring\_peg}: bring the peg to the target peg. \mono{insert\_peg}: insert the peg into the slot.
See \href{https://youtu.be/ESP-x6taBSw}{this video} for solutions of insertion tasks.

}}
{\includegraphics[width=2.5cm]{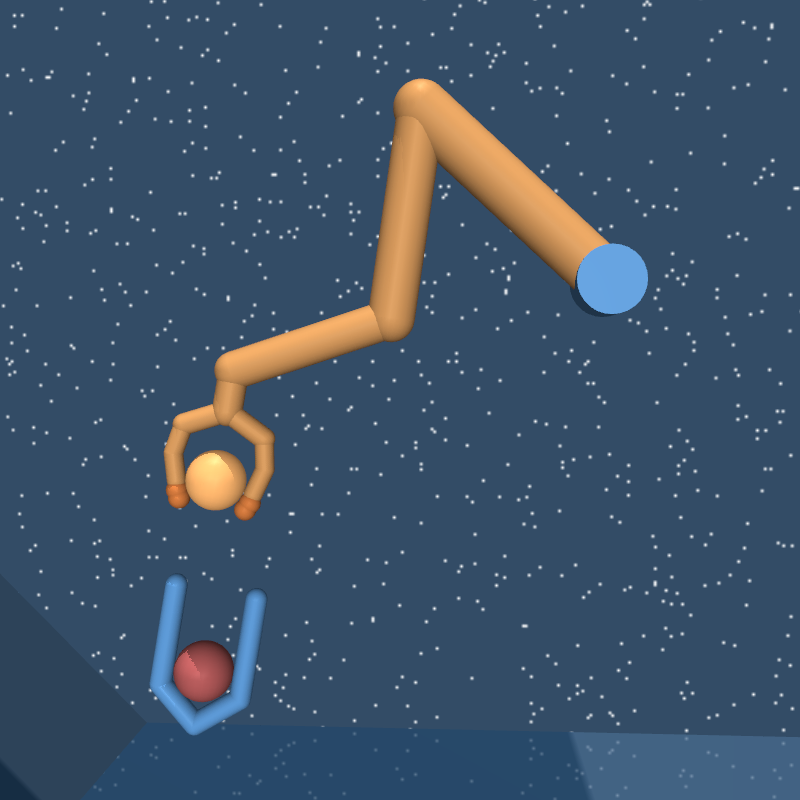}
\includegraphics[width=2.5cm]{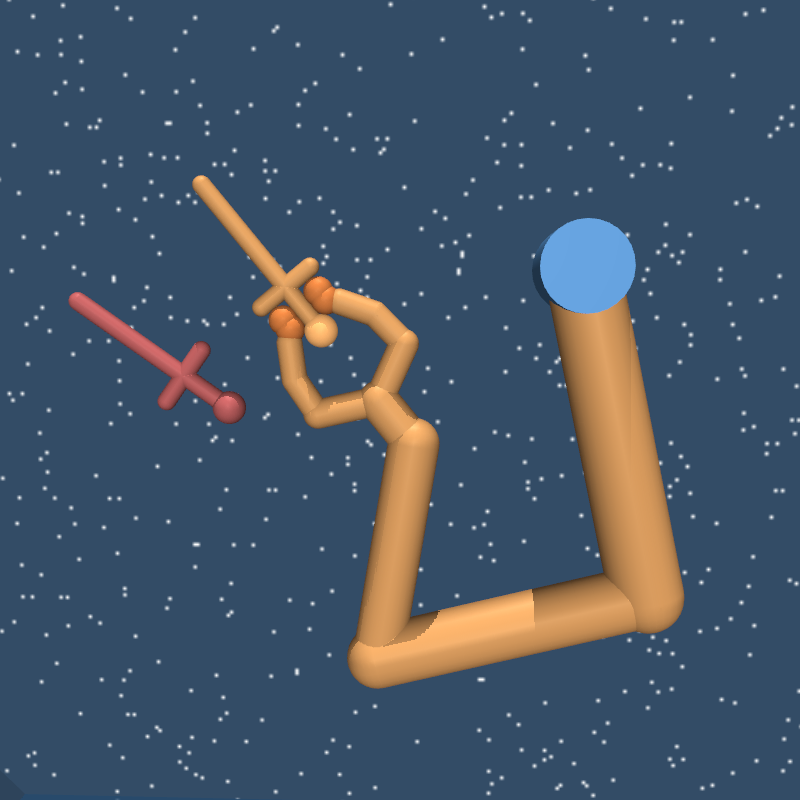}
\includegraphics[width=2.5cm]{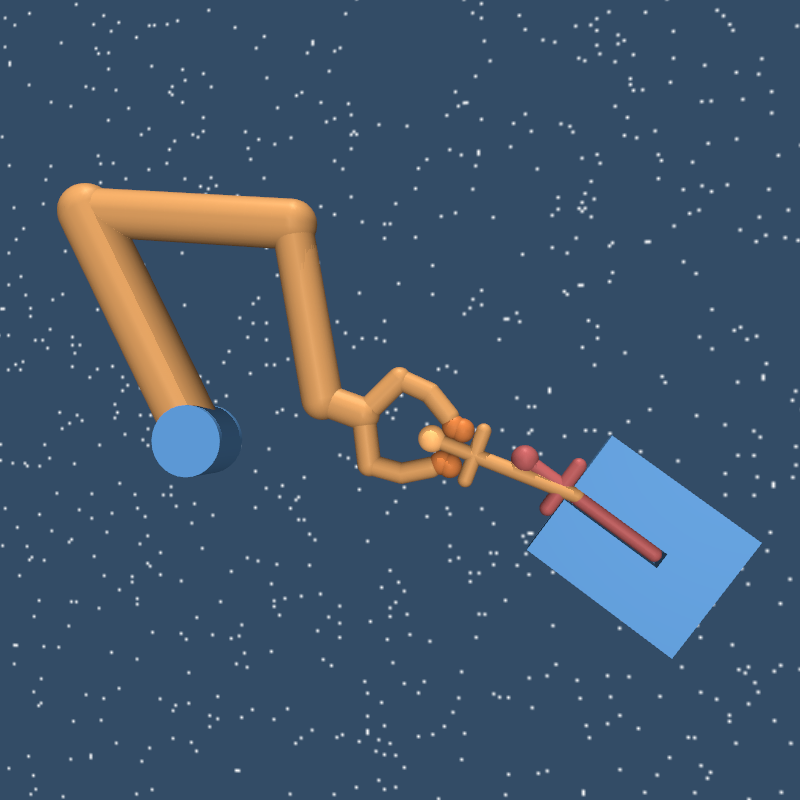}}
\end{figure}
\vspace{-.5cm}

\myfigure{
\textbf{Stacker ($\mathbf{6k\!+\!16,\; 5,\; 11k\!+\!26}$):} Stack $k$ boxes. Reward is given when a box is at the target and the gripper is away from the target, making stacking necessary. The height of the target is sampled uniformly from $\{1,\ldots,k\}$.
}{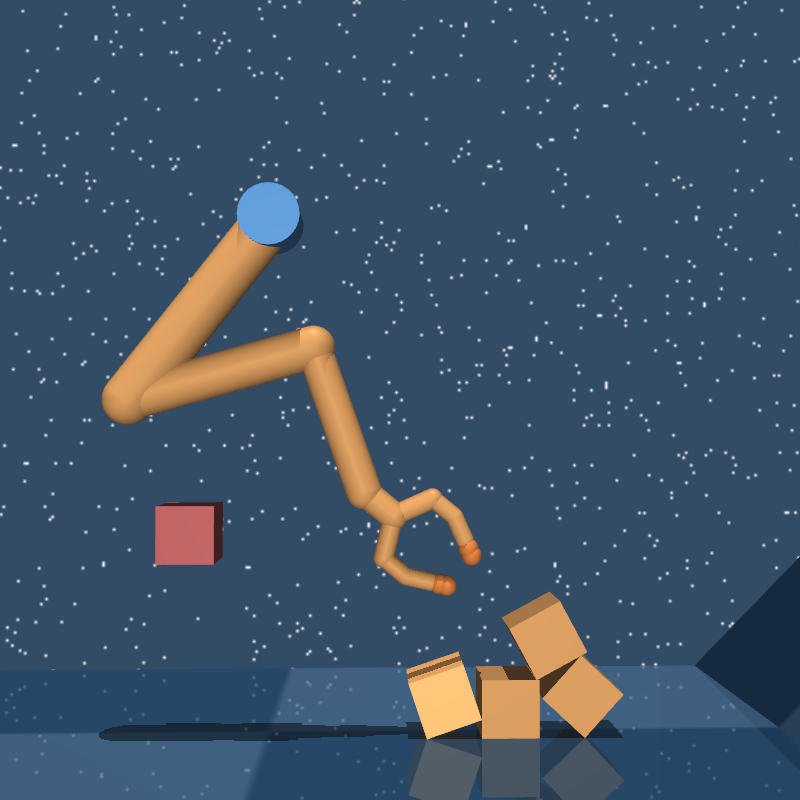}

\myfigure{
\textbf{Swimmer ($\mathbf{2k\!+\!4,\; k\!-\!1,\; 4k\!+\!1}$):}
This procedurally generated $k$-link planar swimmer is based on \citealt{coulom2002reinforcement}, but using \textsf{MuJoCo}'s high-Reynolds fluid drag model. A reward of 1 is provided when the nose is inside the target and decreases smoothly with distance like a Lorentzian. The two instances provided in the \mono{benchmarking} set are the 6-link and 15-link swimmers.
}{./figures/swimmers.png}

\myfigure{
\textbf{Humanoid (54, 21, 67):} A simplified humanoid with 21 joints, based on the model in~\citep{tassa2012synthesis}. Three tasks: \mono{stand}, \mono{walk} and \mono{run} are differentiated by the desired horizontal speed of 0, 1 and $10 \textrm{m/s}$, respectively. Observations are in an egocentric frame and many movement styles are possible solutions e.g.\ running backwards or sideways. This facilitates exploration of local optima.
}{./figures/humanoid.png}

\myfigure{
\textbf{Humanoid\_CMU (124, 56, 137):} A humanoid body with 56 joints, adapted from \citep{merel2017learning} and based on the ASF model of subject~\#8 in the \citetalias{cmu_mocap}. This domain has the same \mono{stand}, \mono{walk} and \mono{run} tasks as the simpler humanoid. We include tools for parsing and playback of the CMU MoCap data, see below. A newer version of this model is now available; see Section \ref{sec:locomotion}.
}{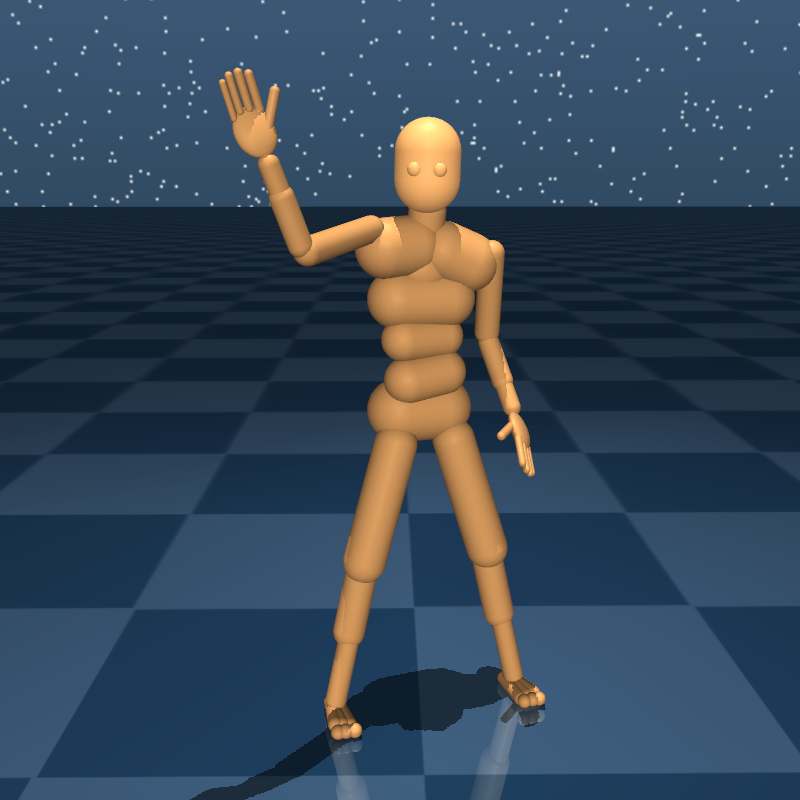}

\myfigure{
\textbf{LQR (2n, m, 2n):} $n$ masses, of which $m$ ($\leq n$) are actuated, move on linear joints which are connected serially. The reward is a quadratic in the position and controls. Analytic transition and control-gain matrices are derived and the optimal policy and value functions are computed in \mono{lqr\_solver.py} using Riccati iterations.
Since both controls and reward are unbounded, \mono{LQR} is not in the \mono{benchmarking} set. 
}{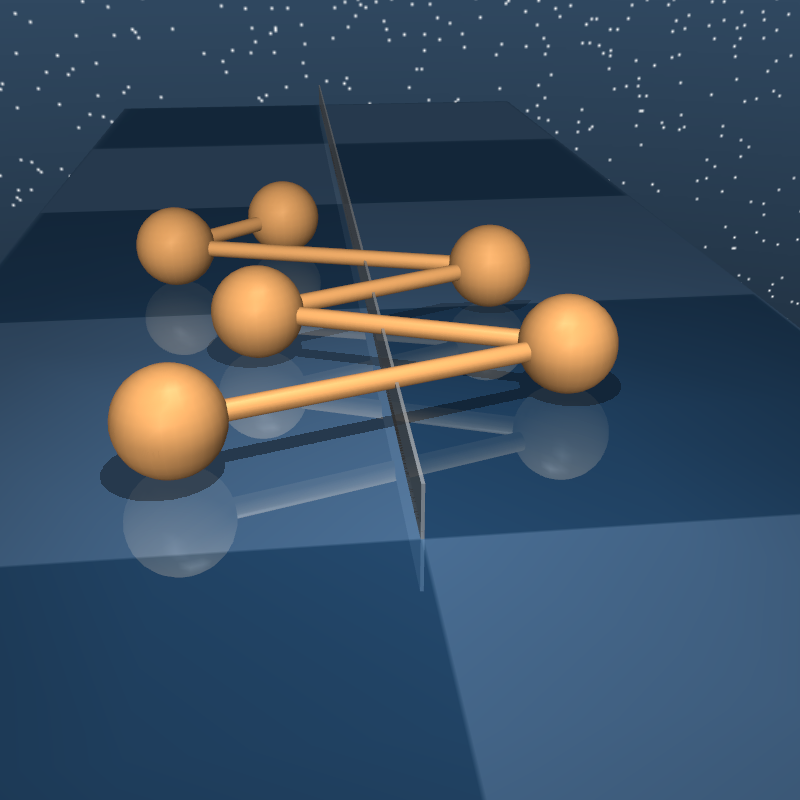}

\subsubsection*{Control Suite Benchmarking}

Please see the original tech report for the \textsf{Control Suite}~\citep{deepmindcontrolsuite2018} for detailed benchmarking results of the \mono{BENCHMARKING} tasks, with several popular Reinforcement Learning algorithms.

\subsection{Additional domains}

\subsubsection*{CMU Motion Capture Data}
We enable \mono{humanoid\_CMU} to be used for imitation learning as in \cite{merel2017learning}, by providing tools for parsing, conversion and playback of human motion capture data from the \citetalias{cmu_mocap}. The \mono{convert()} function in the \mono{parse\_amc} module loads an AMC data file and returns a sequence of configurations for the \mono{humanoid\_CMU} model. The example script \mono{CMU\_mocap\_demo.py} uses this function to generate a video.

\subsubsection*{Quadruped (56, 12, 58)}

\begin{figure*}[ht]
\centering
\begin{minipage}[c]{1\textwidth}
\includegraphics[width=0.32\textwidth]{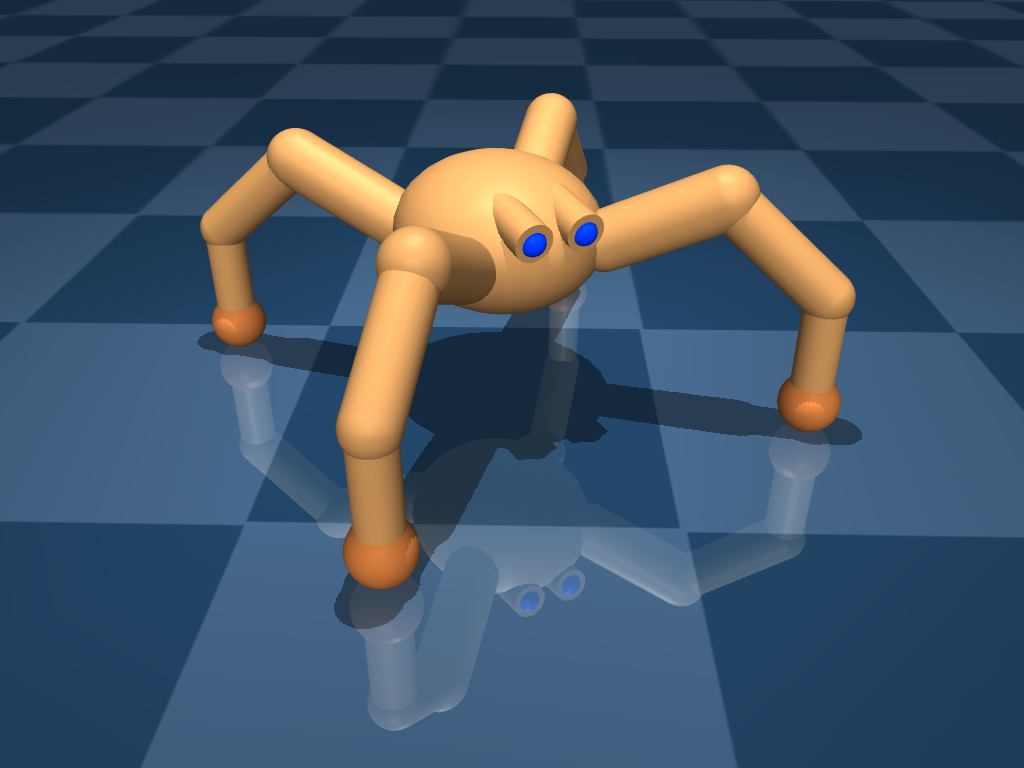}
\hspace{.001\textwidth}
\includegraphics[width=0.32\textwidth]{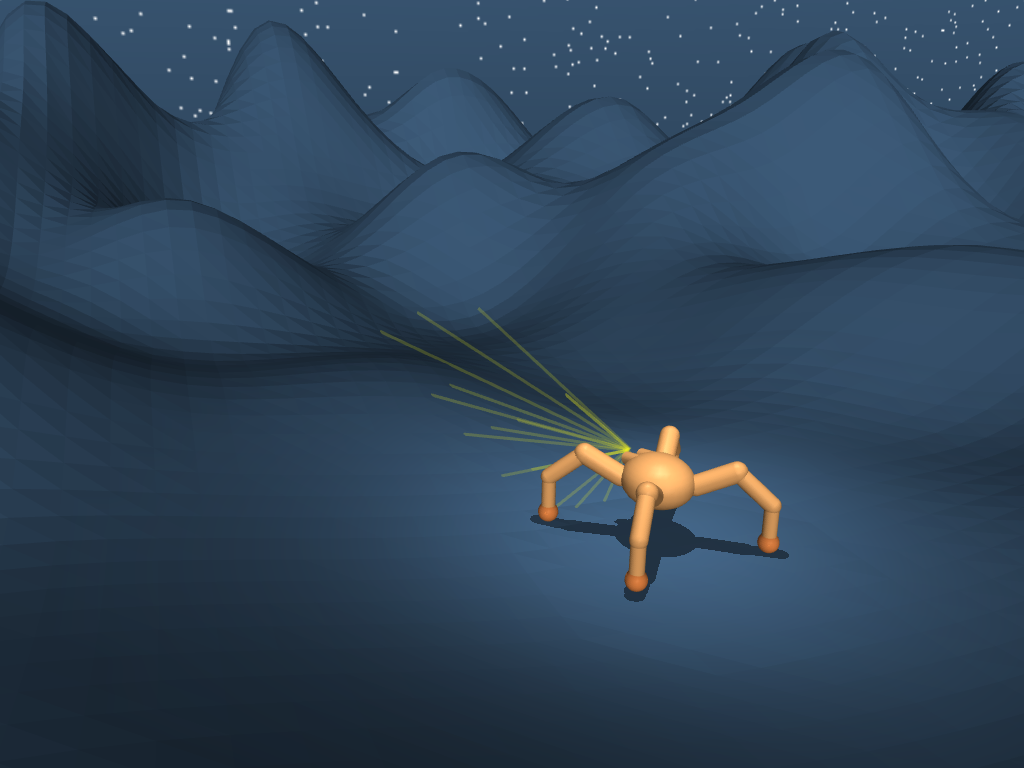}
\hspace{.001\textwidth}
\includegraphics[width=0.32\textwidth]{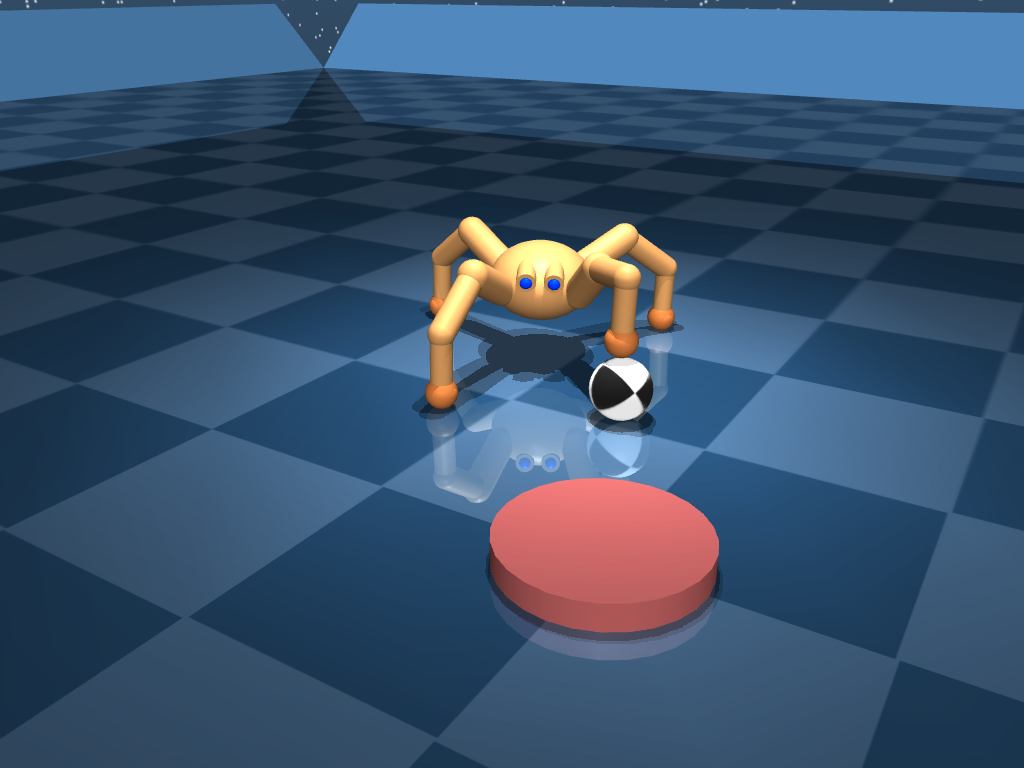}
\caption{\label{fig:quad}\emph{Left:} The Quadruped. \emph{Middle:} In the \mono{escape} task the Quadruped must escape from random mountainous terrain using its rangefinder sensors. \emph{Right:} In the \mono{fetch} task the quadruped must fetch a moving ball and bring it to the red target.}
\end{minipage}
\end{figure*}

The quadruped (Figure \ref{fig:quad}) has 56 state dimensions. Each leg has 3 actuators for a total of 12 actions. Besides the basic \mono{walk} and \mono{run} tasks on flat ground, in the \mono{escape} task the quadruped must climb over procedural random terrain using an array of 20 range-finder sensors (Figure \ref{fig:quad}, \textit{middle}). In the \mono{fetch} task (Figure~\ref{fig:quad}, \textit{right}), the quadruped must run after a moving ball and dribble it to a target at the centre of an enclosed arena. See solutions in \href{https://youtu.be/RhRLjbb7pBE}{\textsf{youtu.be/RhRLjbb7pBE}}.
\vspace{-.5cm}
\subsubsection*{Dog (158, 38, 227)}
\begin{figure*}[htbp]
\centering
\begin{minipage}[c]{1\textwidth}
\includegraphics[height=0.21\textwidth]{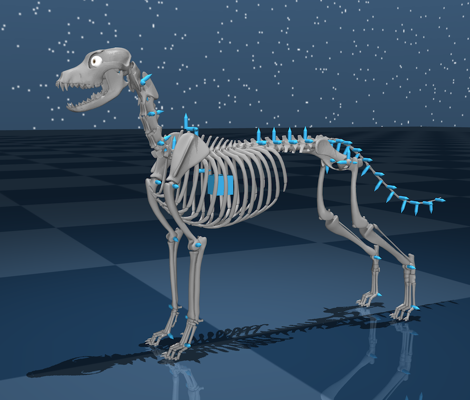}
\hspace{.001\textwidth}
\includegraphics[height=0.21\textwidth]{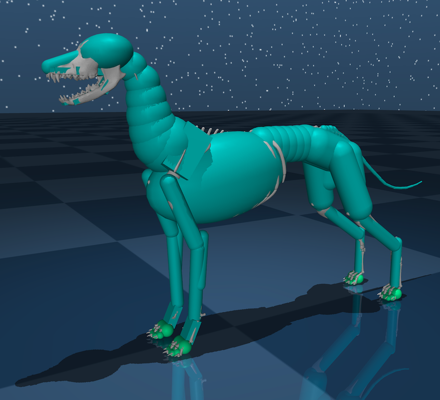}
\hspace{.001\textwidth}
\includegraphics[height=0.21\textwidth]{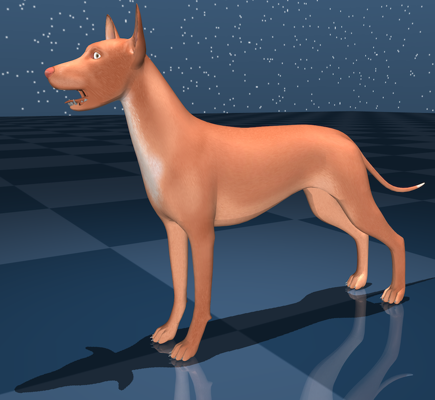}
\hspace{.001\textwidth}
\includegraphics[height=0.21\textwidth]{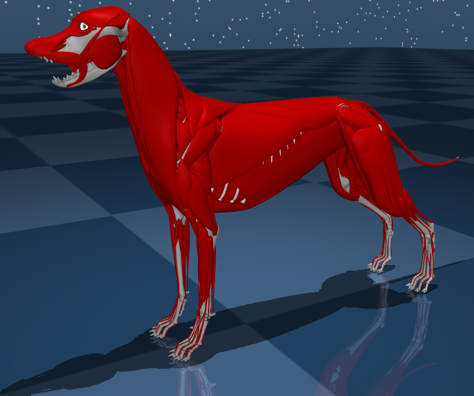}
\caption{\label{fig:dog}\emph{Left:} Dog skeleton, joints visualised as light blue elements. \emph{Middle-Left:} Collision geometry, overlain with skeleton. \emph{Middle-Right:} Textured skin, used for visualisation only. \emph{Right:} Dog muscles, included but not rigged to the skeleton. See \href{https://youtu.be/i0_OjDil0Fg}{\textsf{youtu.be/i0\_OjDil0Fg}} for preliminary solution of the \mono{run} and \mono{fetch} tasks.}
\end{minipage}
\end{figure*}

\noindent A realistic model of a Pharaoh Dog (Figure \ref{fig:dog}) was prepared for DeepMind by \href{https://www.turbosquid.com/Search/Artists/leo3Dmodels}{leo3Dmodels} and is made available to the wider research community. The kinematics, skinning weights and collision geometry are created procedurally using \textsf{PyMJCF}. The static model includes muscles and tendon attachment points (Figure~\ref{fig:dog}, \emph{Right}). Including these in the dynamical model using \textsf{MuJoCo}'s support for tendons and muscles requires detailed anatomical knowledge and remains future work.

\subsubsection*{Rodent (184, 38, 107 + 64$\times$64$\times$3 pixels)}
In order to better compare learned behaviour with experimental settings common in the life sciences, we have built a model of a rodent.  See \citep{merel2020deep} for initial research training a policy to control this model using visual inputs and analysing the resulting neural representations. Related videos of rodent tasks therein:  \href{https://www.youtube.com/watch?v=vBIV1qJpJK8}{``forage''}, \href{https://www.youtube.com/watch?v=rFelC_YbeLE}{``gaps''},
\href{https://www.youtube.com/watch?v=6d0SX56Cn6Q}{``escape''} and
\href{https://www.youtube.com/watch?v=lBKwHzO-z_0}{``two-tap''}. The skeleton reference model was made by \href{https://www.turbosquid.com/Search/Artists/leo3Dmodels}{leo3Dmodels} (not included).  

\begin{figure*}[ht]
\centering
\begin{minipage}[c]{1\textwidth}
\includegraphics[height=0.175\textwidth]{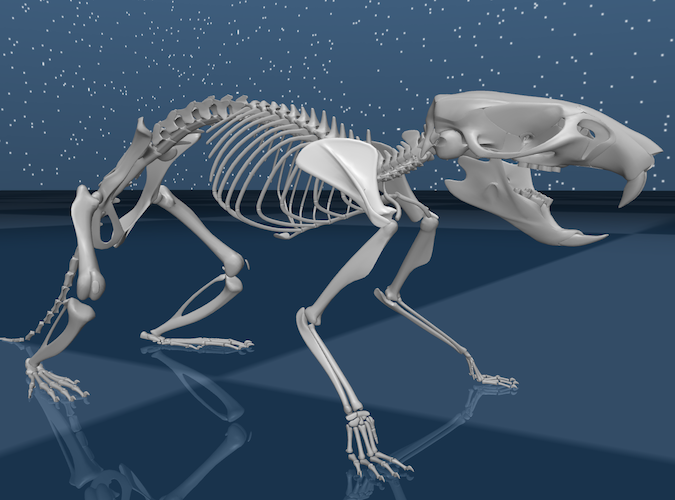}
\hspace{.0001\textwidth}
\includegraphics[height=0.175\textwidth]{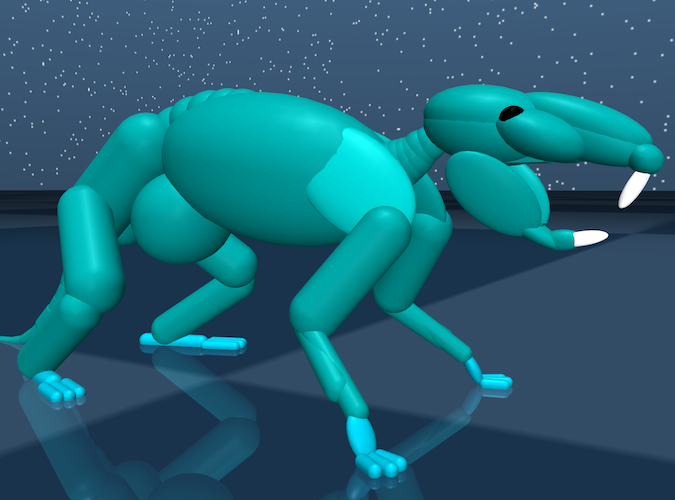}
\hspace{.0001\textwidth}
\includegraphics[height=0.175\textwidth]{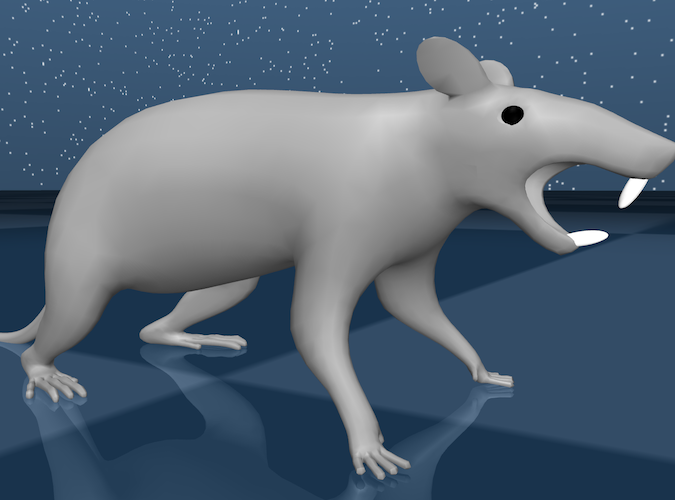}
\hspace{.0001\textwidth}
\includegraphics[height=0.175\textwidth]{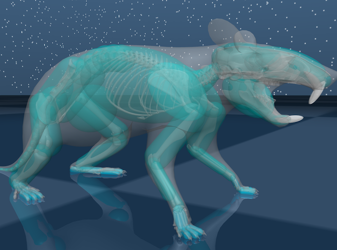}
\caption{
\label{fig:rat} Rodent; figure reproduced from \citealt{merel2020deep}.
\emph{Left}: Anatomical skeleton of a rodent (as reference; not part of physical simulation). \emph{Middle-Left}: Collision geometry designed around the skeleton.
\emph{Middle-Right}: Cosmetic skin to cover the body. \emph{Right}: Semi-transparent visualisation of the three layers overlain.
}
\end{minipage}
\end{figure*}

\section{Locomotion tasks}
\label{sec:locomotion}
Inspired by our early work in \citealt{heess2017emergence}, the \textsf{Locomotion} library provides a framework and a set of high-level components for creating rich locomotion-related task domains. The central abstractions are the Walker, an agent-controlled \textsf{Composer} Entity that can move itself, and the Arena, the physical environment in which behaviour takes place. Walkers expose locomotion-specific methods, like observation transformations into an egocentric frame, while Arenas can re-scale themselves to fit Walkers of different sizes. Together with the Task, which includes a specification of episode initialisation, termination, and reward logic, a full RL environment is specified. The library currently includes navigating a corridor with obstacles, foraging for rewards in a maze, traversing rough terrain and multi-agent soccer. Many of these tasks were first introduced in \citealt{merel2018hierarchical} and \citealt{liu2019emergent}.

\subsection{Humanoid running along corridor with obstacles}
\label{sec:locomotion_walls}
\begin{figure}[ht]
    \centering
    \includegraphics[width=1.0\textwidth]{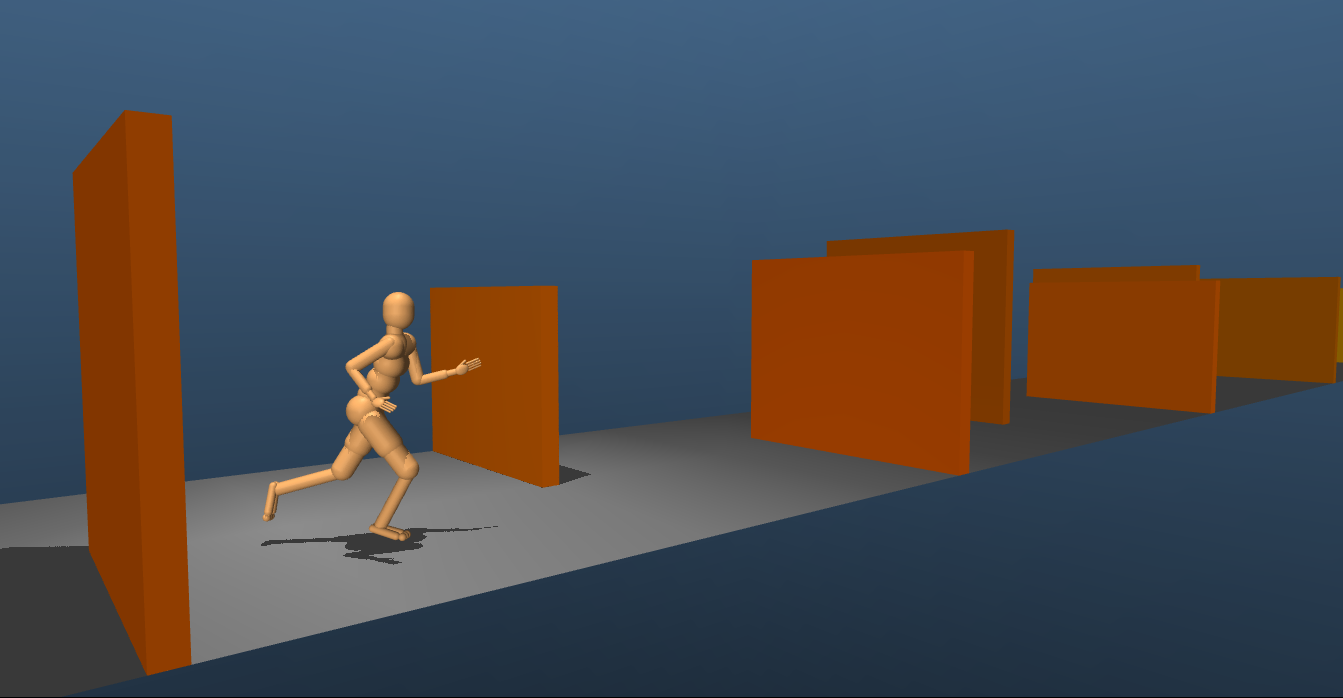}
    \caption{A perspective of the environment in which the humanoid is tasked with navigating around walls along a corridor.}
    \label{fig:locomotion_walls}
\end{figure}
As an illustrative example of using the \textsf{Locomotion} infrastructure to build an RL environment, consider placing a humanoid in a corridor with walls, and a task specifying that the humanoid will be rewarded for running along this corridor, navigating around the wall obstacles using vision. We instantiate the environment as a composition of the Walker, Arena, and Task as follows. First, we build a position-controlled CMU humanoid walker. 
\begin{lstlisting}[language=Python]
walker = cmu_humanoid.CMUHumanoidPositionControlledV2020(
  observable_options={'egocentric_camera': dict(enabled=True)})
\end{lstlisting}
Note that this CMU humanoid is ``\textsf{V2020}'', an improved version from the initial one released in \cite{deepmindcontrolsuite2018}. Modifications include overall body height and mass better matching a typical human, more realistic body proportions, and better tuned gains and torque limits for position-control actuators. 
\newpage
\noindent Next, we construct a corridor-shaped arena that is obstructed by walls.
\begin{lstlisting}[language=Python]
arena = arenas.WallsCorridor(wall_gap=3.,
                             wall_width=distributions.Uniform(2., 3.),
                             wall_height=distributions.Uniform(2.5, 3.5),
                             corridor_width=4.,
                             corridor_length=30.)
\end{lstlisting}
Finally, a task that rewards the agent for running down the corridor at a specific velocity is instantiated as a \mono{composer.Environment}.
\begin{lstlisting}[language=Python]
task = tasks.RunThroughCorridor(walker=walker,
                                arena=arena,
                                walker_spawn_position=(0.5, 0, 0),
                                target_velocity=3.0,
                                physics_timestep=0.005,
                                control_timestep=0.03)

environment = composer.Environment(time_limit=10,
                                   task=task,
                                   strip_singleton_obs_buffer_dim=True)
\end{lstlisting}
 \href{https://youtu.be/UfSHdOg-bOA}{\texttt{youtu.be/UfSHdOg-bOA}} shows a video of a solution of this task, produced with \citealt{abdolmaleki2018maximum}'s MPO agent.

\subsection{Maze navigation and foraging}
\label{sec:maze}
\vspace{-.3cm}
\begin{figure}[H]
\floatbox[{\capbeside\thisfloatsetup{capbesideposition={left,top},
capbesidewidth=0.57\textwidth}}]{figure}[\FBwidth]
{\caption*{\normalsize
We include a procedural maze generator for arenas (the same one used in \citealt{dmlab2016}), to construct navigation and foraging tasks. On the right is the CMU humanoid in a human-sized maze, with spherical rewarding elements. \href{https://www.youtube.com/watch?v=vBIV1qJpJK8}{youtu.be/vBIV1qJpJK8} shows the Rodent navigating a rodent-scale maze arena.
}}
{\includegraphics[width=0.4\textwidth]{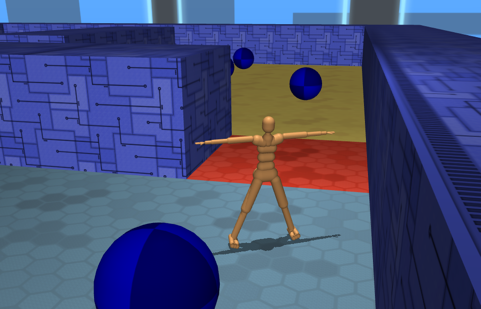}}
\end{figure}
\subsection{Multi-Agent soccer}

\begin{figure*}[ht]
\centering
\includegraphics[height=0.365\textwidth]{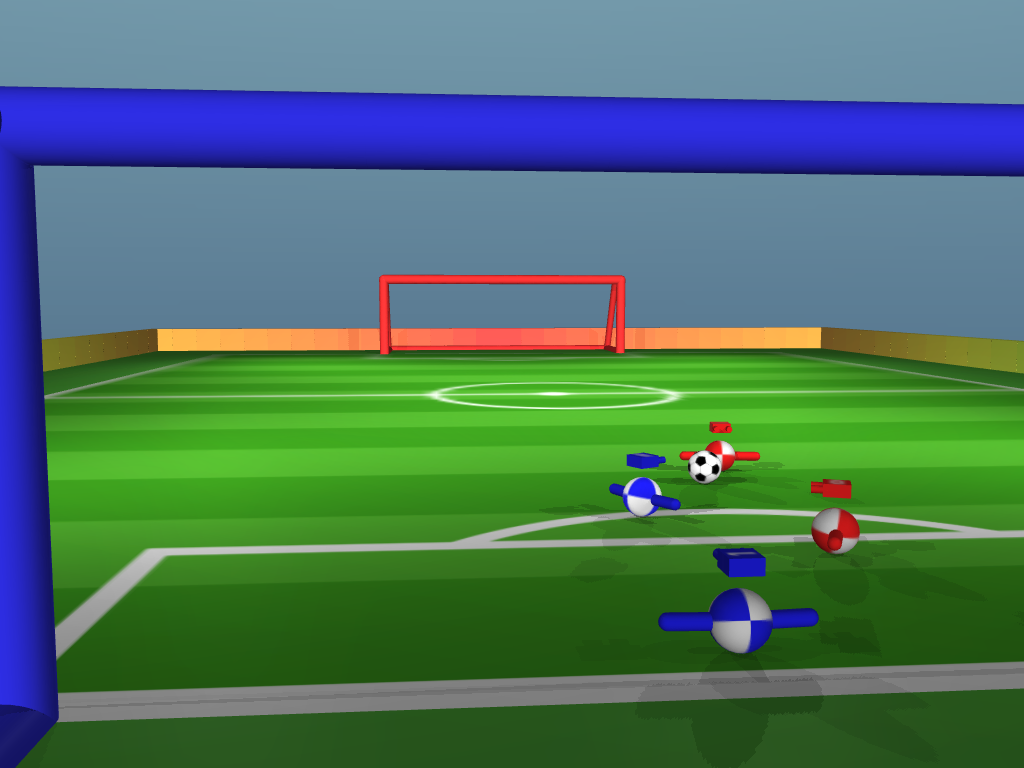}
\hspace{.001\textwidth}
\includegraphics[height=0.365\textwidth]{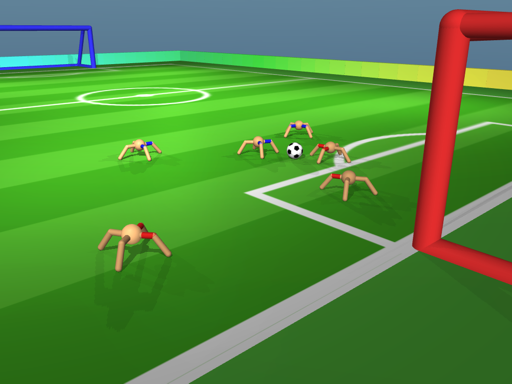}
\caption{\label{fig:locomotion_soccer} Rendered scenes of \mono{Locomotion} multi-agent soccer \emph{Left:} 2-vs-2 with \mono{BoxHead} walkers. \emph{Right:} 3-vs-3 with \mono{Ant}.}
\end{figure*}
\noindent Building on \textsf{Composer} and \textsf{Locomotion} libraries, the \textsf{Multi-agent soccer} environments, introduced in \citealt{liu2019emergent}, follow a consistent task structure of Walkers, Arena, and Task where instead of a single walker, we inject multiple walkers that can interact with each other physically in the same scene. The code snippet below shows how to instantiate a 2-vs-2 \textsf{Multi-agent Soccer} environment with the simple, 5 degree-of-freedom \mono{BoxHead} walker type. For example it can trivially be replaced by \mono{WalkerType.ANT}, as shown in Figure \ref{fig:locomotion_soccer}.

\begin{lstlisting}[language=Python]
from dm_control.locomotion import soccer

team_size = 2
num_walkers = 2 * team_size

env = soccer.load(team_size=team_size,
                  time_limit=45,
                  walker_type=soccer.WalkerType.BOXHEAD)
\end{lstlisting}
To implement a synchronous multi-agent environment, we adopt the convention that each \mono{TimeStep} contains a sequence of per-agent observation dictionaries and expects a sequence of per-agent action arrays in return.
\begin{lstlisting}[language=Python]
assert len(env.action_spec()) == num_walkers
assert len(env.observation_spec()) == num_walkers

# Reset and initialize the environment.
timestep = env.reset()

# Generates a random action according to the `action_spec`.
random_actions = [spec.generate_value() for spec in env.action_spec()]
timestep = env.step(random_actions)

# Check that timestep respects multi-agent action and observation convention.
assert len(timestep.observation) == num_walkers
assert len(timestep.reward) == num_walkers
\end{lstlisting}

\section{Manipulation tasks}
\label{sec:manipulation}

\begin{figure}
    \centering
    \includegraphics[width=\textwidth]{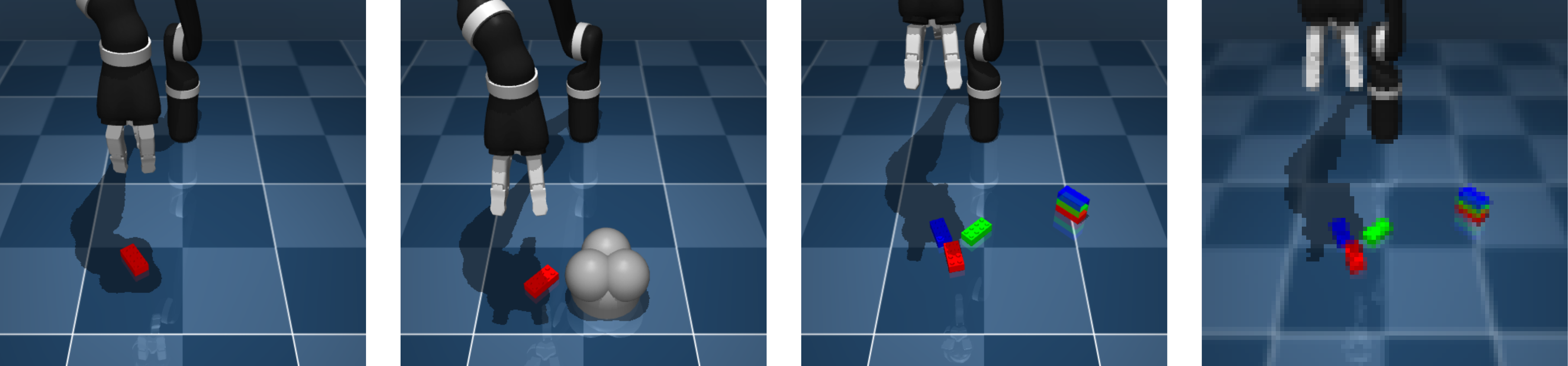}
    \caption{
    Randomly sampled initial configurations for the \mono{lift\_brick}, \mono{place\_cradle}, and \mono{stack\_3\_bricks} environments (left to right).
    The rightmost panel shows the corresponding 84x84 pixel visual observation returned by \mono{stack\_3\_bricks\_vision}.
    Note the stack of three translucent bricks to the right of the workspace, representing the goal configuration.
    }
    \label{fig:manipulation_tasks}
\end{figure}

The \textsf{manipulation} module provides a robotic arm, a set of simple objects, and tools for building reward functions for manipulation tasks.
Each example environment comes in two different versions that differ in the types of observation available to the agent:

\begin{enumerate}[parsep=0cm]
    \item \mono{features}
    \begin{itemize}
        \item Arm joint positions, velocities, and torques.
        \item Task-specific privileged features (including the positions and velocities of other movable objects in the scene).
    \end{itemize}
    \item \mono{vision}
    \begin{itemize}
        \item Arm joint positions, velocities, and torques.
        \item Fixed RGB camera view showing the workspace.
    \end{itemize}
\end{enumerate}
All of the manipulation environments return a reward $r(\mathbf{s}, \mathbf{a}) \in [0, 1]$ per timestep, and have an episode time limit of 10 seconds.
The following code snippet shows how to import the manipulation tasks and view all of the available  environments:

\begin{lstlisting}[language=Python]
from dm_control import manipulation

# `ALL` is a tuple containing the names of all of the environments.
print('\n'.join(manipulation.ALL))
\end{lstlisting}
Environments are also tagged according to what types of observation they return.
\mono{get\_environments\_by\_tag} lists the names of environments with specific tags:

\begin{lstlisting}[language=Python]
print('\n'.join(manipulation.get_environments_by_tag('vision')))
\end{lstlisting}
Environments are instantiated by name using the \mono{load} method, which also takes an optional \mono{seed} argument that can be used to seed the random number generator used by the environment.
\begin{lstlisting}[language=Python]
env = manipulation.load('stack_3_bricks_vision', seed=42)
\end{lstlisting}
\subsection{Studded brick model}
\vspace{-.3cm}
\begin{figure}[H]
\floatbox[{\capbeside\thisfloatsetup{capbesideposition={left,top},
capbesidewidth=0.67\textwidth}}]{figure}[\FBwidth]
{\caption*{\normalsize
The \mono{stack\_3\_bricks\_vision} task, like most of the included manipulation tasks, makes use of the studded bricks shown on the right. These were modelled on Lego Duplo{\textregistered} bricks, snapping together when properly aligned and force is applied, and holding together using friction.
}}
{\includegraphics[width=0.3\textwidth]{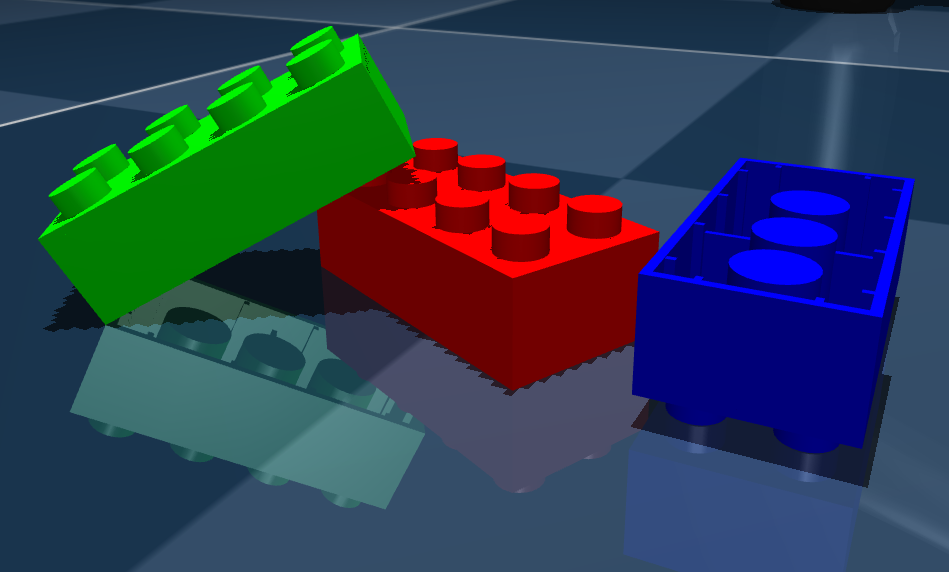}}
\end{figure}
\vspace{-.2cm}
\subsection{Task descriptions}
Brief descriptions of each task are given below:
\begin{itemize}[parsep=0cm]
    \item \mono{reach\_site}:
    Move the end effector to a target location in 3D space.

    \item \mono{reach\_brick}:
    Move the end effector to a brick resting on the ground.

    \item \mono{lift\_brick}:
    Elevate a brick above a threshold height.
    
    \item \mono{lift\_large\_box}:
    Elevate a large box above a threshold height.
    The box is too large to be grasped by the gripper, requiring non-prehensile  manipulation.

    \item \mono{place\_cradle}:
    Place a brick inside a concave `cradle' situated on a pedestal.

    \item \mono{place\_brick}:
    Place a brick on top of another brick that is attached to the top of a pedestal.
    Unlike the stacking tasks below, the two bricks are not required to be snapped together in order to obtain maximum reward.

    \item \mono{stack\_2\_bricks}:
    Snap together two bricks, one of which is attached to the floor.

    \item \mono{stack\_2\_bricks\_moveable\_base}:
    Same as \mono{stack\_2\_bricks}, except both bricks are movable.

    \item \mono{stack\_2\_of\_3\_bricks\_random\_order}:
    Same as \mono{stack\_2\_bricks}, except there is a choice of two color-coded movable bricks, and the agent must place the correct one on top of the fixed bottom brick.
    The goal configuration is represented by a visual hint consisting of a stack of translucent, contactless bricks to the side of the workspace.
    In the \mono{features} version of the task the observations also contain a vector of indices representing the desired order of the bricks.

    \item \mono{stack\_3\_bricks}:
    Assemble a tower of 3 bricks.
    The bottom brick is attached to the floor, whereas the other two are movable.
    The top two bricks must be assembled in a specific order.

    \item \mono{stack\_3\_bricks\_random\_order}:
    Same as \mono{stack\_3\_bricks}, except the order of the top two bricks does not matter.

    \item \mono{reassemble\_3\_bricks\_fixed\_order}:
    The episode begins with all three bricks already assembled in a stack, with the bottom brick being attached to the floor.
    The agent must disassemble the top two bricks in the stack, and reassemble them in the opposite order.

    \item \mono{reassemble\_5\_bricks\_random\_order}:
    Same as the previous task, except there are 5 bricks in the initial stack.
    There are therefore $4!-1$ possible alternative configurations in which the top 4 bricks in the stack can be reassembled, of which only one is correct.

\end{itemize}

\section{Conclusion}

\dmcontrol is a starting place for the testing and  performance comparison of reinforcement learning algorithms for physics-based control. It offers a wide range of pre-designed RL tasks and a rich framework for designing new ones. We are excited to be sharing these tools with the wider community and hope that they will be found useful. We look forward to the diverse research the Control Suite and associated libraries may enable, and to integrating community contributions in future releases.

\newpage

\section{Acknowledgements}
We would like to thank Raia Hadsell, Yori Zwols and Joseph Modayil for their reviews; Yazhe Li and Diego de Las Casas for their help with the Control Suite; Ali Eslami and Guy Lever for their contributions to the soccer environment.


{\small
\bibliographystyle{plainnat}
\bibliography{main}
}

\end{document}